\documentclass[twocolumn]{article}
\usepackage{PRIMEarxiv}

\usepackage[utf8]{inputenc} 
\usepackage[T1]{fontenc}    
\usepackage{hyperref}       
\usepackage{url}            
\usepackage{booktabs}       
\usepackage{amsfonts}       
\usepackage{nicefrac}       
\usepackage{microtype}      
\usepackage{lipsum}
\usepackage{fancyhdr}       
\usepackage{graphicx}       
\graphicspath{{media/}}     
\usepackage{hyperref}
\usepackage{enumitem}
\usepackage{url}
\usepackage{xcolor}
\usepackage{amsmath}

\usepackage{tabularx}
\usepackage{booktabs}

\usepackage{epstopdf}

\newcolumntype{n}[1]{>{\hsize=#1\hsize\raggedright\arraybackslash}X}

\newcommand{\subcaption}[1]{%
  \textcolor{black}{\footnotesize(#1)}\quad
}

\newcommand{\bottomsubcaptionsABC}{%
  \vspace{0ex}
  \par\noindent
  \small
  \textcolor{black}{%
    \begin{tabular}{@{}p{0.33\linewidth}@{\hskip 0pt}p{0.33\linewidth}@{\hskip 0pt}p{0.33\linewidth}@{}}
      \centering (a) &
      \centering (b) &
      \centering (c)
    \end{tabular}%
  }
}

\newcounter{biography}
\makeatletter
\newenvironment{myBiographyNoPhoto}[1]{%
  \stepcounter{biography}%
  \normalfont\interlinepenalty500%
  \vskip 1\baselineskip plus 1fil minus 0\baselineskip
  \parskip=0pt\par%
  \noindent{\textbf{\MakeUppercase{#1}\ }}\ignorespaces}{\relax\par\normalfont}
\makeatother

\newcommand{\hl}[1]{#1}
\usepackage{enumitem}
\setlist[itemize]{leftmargin=0.5cm}
\title{RGB-D And Thermal Sensor Fusion: A Systematic Literature Review}

\author{
  Martin Brenner \\
  Massey University \\
  Auckland, New Zealand\\
  \texttt{martin.brenner.1@uni.massey.ac.nz} \\
  \AND
  Napoleon H. Reyes \\
  Massey University \\
  Auckland, New Zealand \\
  \AND
  Teo Susnjak \\
  Massey University \\
  Auckland, New Zealand \\
  \And
  Andre L.C. Barczak \\
  Bond University \\
  Gold Coast, Australia \\
}

\begin{document}
\twocolumn[
  \begin{@twocolumnfalse}
    \maketitle
    \begin{abstract}
    In the last decade, the computer vision field has seen significant progress in multimodal data fusion and learning, where multiple sensors, including depth, infrared, and visual, are used to capture the environment across diverse spectral ranges. Despite these advancements, there has been no systematic and comprehensive evaluation of fusing RGB-D and thermal modalities to date. While autonomous driving using LiDAR, radar, RGB, and other sensors has garnered substantial research interest, along with the fusion of RGB and depth modalities, the integration of thermal cameras and, specifically, the fusion of RGB-D and thermal data, has received comparatively less attention. This might be partly due to the limited number of publicly available datasets for such applications.
    This paper provides a comprehensive review of both, state-of-the-art and traditional methods used in fusing RGB-D and thermal camera data for various applications, such as site inspection, human tracking, fault detection, and others. The reviewed literature has been categorised into technical areas, such as 3D reconstruction, segmentation, object detection, available datasets, and other related topics. Following a brief introduction and an overview of the methodology, the study delves into calibration and registration techniques, then examines thermal visualisation and 3D reconstruction, before discussing the application of classic feature-based techniques and modern deep learning approaches. The paper concludes with a discourse on current limitations and potential future research directions. It is hoped that this survey will serve as a valuable reference for researchers looking to familiarise themselves with the latest advancements and contribute to the RGB-DT research field.
    \end{abstract}
    \vspace{1cm}
  \end{@twocolumnfalse}
]
\section{Introduction}
\label{sec:introduction}
\begin{figure*}[ht]
  \centering
    \includegraphics[width=0.8\textwidth]{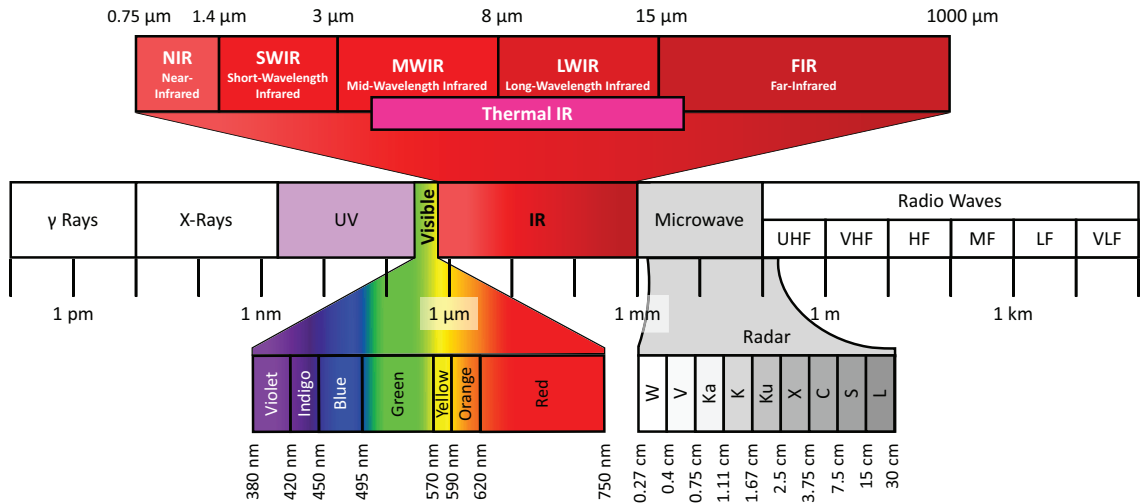}
    \caption{\hl{Depiction of the electromagnetic spectrum, with a focus on the various infrared (IR) bands. The thermal IR range, which is radiation-based, is specifically indicated.}}
    \label{fig:000 spectral_range}
\end{figure*}

\begin{figure}[ht]
    \centering
    \includegraphics[width=0.48\textwidth]{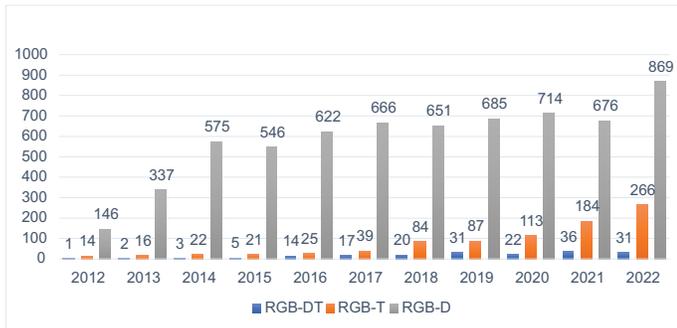}
    \caption{Number of publications with RGB-DT, RGB-T and RGB-D from 2012 to 2022.
Source: Data from Google Scholar keyword search: ("RGB-DT" OR "RGB-DT"); ("rgb-t" OR "rgbt" OR "RGB-Thermal"); ("rgb-d" OR "rgbd" OR "RGB-Depth" OR "RGB+Depth")[peer reviewed articles only]}
    \label{fig:000 NumberOfStudiesRGBDT}
\end{figure}

The extraction and analysis of features from RGB images have become a widely used processing technique in computer vision, finding its way into a diverse array of industrial, commercial, and everyday applications. However, this technique exhibits limitations, primarily from its confinement to the visible spectrum. As illustrated in Fig. \ref{fig:000 spectral_range}, the visible imaging range is notably narrower compared to other spectra, which underscores the potential benefits of exploring alternative non-visible spectral regions to overcome these restrictions.
The most significant constraint is that it only operates effectively under good lighting conditions and clear visibility. This has prompted researchers to explore using RGB-D and thermal cameras for multi-spectral perception in recent years as shown in Fig. \ref{fig:000 NumberOfStudiesRGBDT}. 

The increasing application of depth cameras can largely be attributed to the release of the Microsoft Kinect sensor in 2010. This sensor utilises an infrared (IR) structured light system, operating in the Near Infra Red (NIR) band, to capture depth information in addition to RGB colour data, and was the first depth camera to be widely available for the consumer market. Thermal cameras on the other hand capture temperature information and have been available for many years. Despite a drop in price, the cost of thermal cameras is still considerably high and the possible resolution of the sensors is low compared to RGB cameras due to the larger pixel pitch required for the Long-Wave Infrared(LWIR) band. With the introduction of the first microbolometric array camera in 1997, detector cooling in thermal cameras became unnecessary \cite{RN114}, as non-cooled thermal imagers now feature electronic stabilisation. While non-cooled cameras offer advantages such as being lighter, faster, more affordable, and more reliable, cooled cameras still have the edge in terms of greater sensitivity \cite{RN101}. 

\hl{In the context of sensor fusion, each sensor modality - RGB, Depth (D), and Thermal (T) - brings its own set of advantages and disadvantages.}

\hl{RGB cameras, being ubiquitous and economical, offer high-resolution colour images that are readily interpreted by both human observers and computer vision algorithms. They excel in tasks such as object recognition, scene understanding, and texture analysis. However, their efficacy is heavily dependent on good lighting conditions.}

\hl{Conversely, depth sensors, integral to RGB-D cameras, operate relatively independently of visible light, allowing them to function effectively under a variety of lighting conditions. Nevertheless, their range is typically limited, and they can be affected by factors such as sunlight interference (in the case of Time of Flight sensors) or low texture areas and lighting conditions (in the case of stereo vision). Despite these limitations, depth sensors offer valuable 3D environmental information, proving advantageous for tasks such as object detection, localisation, and navigation.}

\hl{Thermal infrared sensors, in contrast to the visible and depth modalities, can sense slight temperature differences between objects and their surroundings. This capability is not hindered by low-light conditions or complete darkness, as these sensors operate based on thermal radiation, independent of any light source. This unique capability makes thermal sensing a valuable modality for object detection under challenging conditions. However, thermal sensors typically offer lower resolution than RGB cameras and are more expensive.}

\hl{The overarching aim of sensor fusion is to harmonise the strengths of each sensor modality to mitigate their individual limitations. However, achieving effective fusion requires careful calibration and alignment of the sensors, along with sophisticated algorithms to integrate the different types of data.}

The field of surveillance has shown significant interest in the integration of RGB and Thermal (RGB-T) data. Similarly, the combination of LiDAR sensors or stereo depth cameras with RGB, polarised images, and radar is a well-explored area in autonomous vehicles and robotics. However, the fusion of RGB-D and thermal data has not been studied as extensively in comparison. 
Fusion of these three modalities has the potential to provide more robust and accurate perception in various applications, such as object recognition, tracking, and localisation for applications where no long-range detection is required or the detection of endotherms is beneficial. 

LiDAR and RGB-D cameras are both used for capturing 3D data, but they have different characteristics. LiDAR produces a sparser 3D point cloud with decreasing resolution over distance, while RGB-D cameras produce a more densely packed depth map that is limited to a few metres of distance. RGB-D sensors that rely on IR Time Of Flight (ToF) technology are not suitable for outdoor applications due to interference from sunlight, but devices based on stereo vision can overcome this issue. 

In order to achieve an effective fusion of the different modalities, it is essential to calibrate each sensor and align them in the same coordinate system, which involves determining the intrinsic and extrinsic parameters using the pin-hole camera model. Aligning the modalities correctly is crucial for achieving precise data fusion. Although descriptor-based methods utilising feature point matching algorithms can accomplish registration, they are often not suitable for real-time applications involving moving cameras. This is due to their high computational complexity and the challenges in implementing them with thermal data, which displays distinct characteristics compared to visual data.

\begin{figure}[ht]
    \centering
    \includegraphics[width=0.48\textwidth]{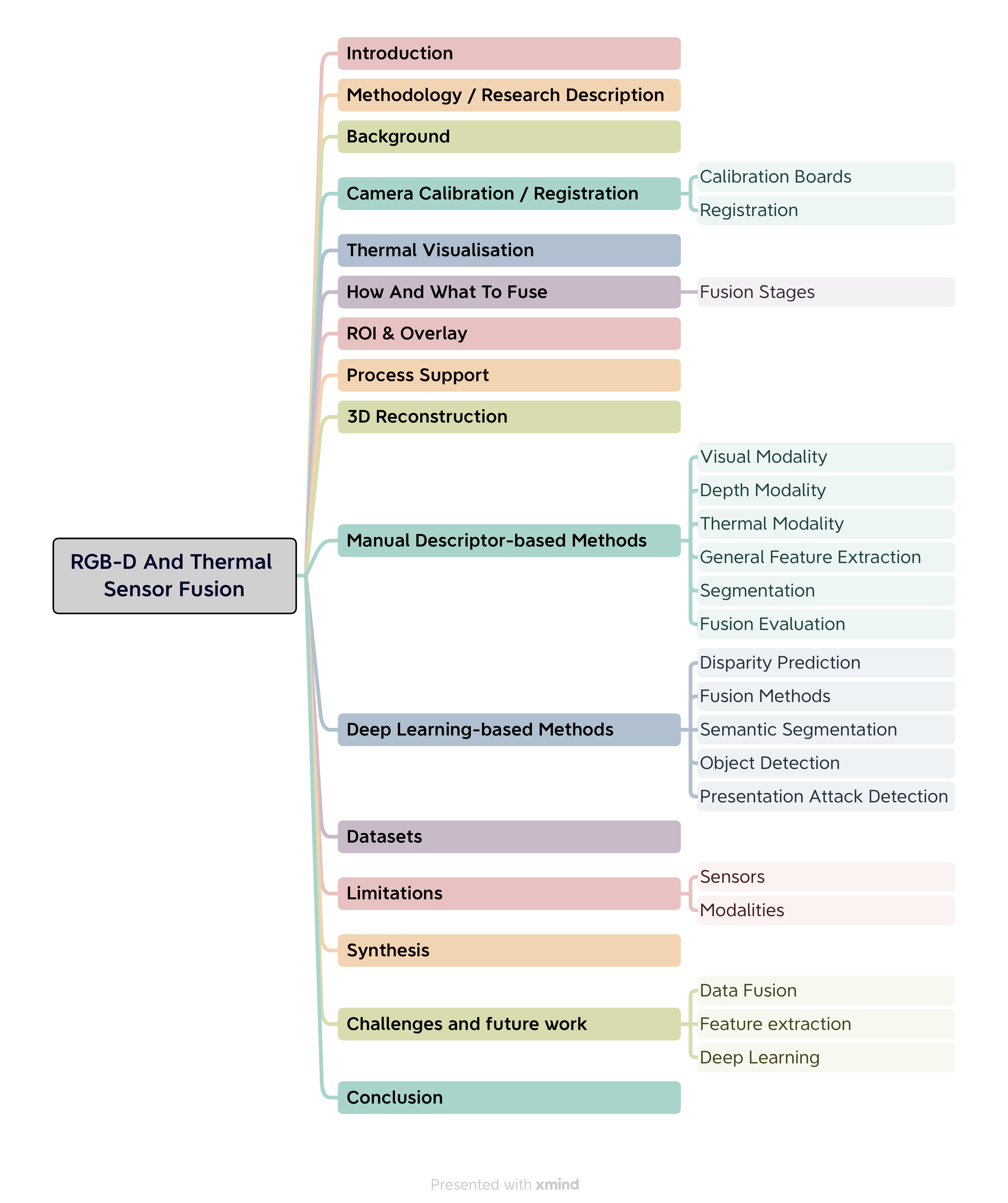}
    \caption{The overall structure of this paper.}
    \label{fig:000 paper_structure}
\end{figure}
\begin{figure*}[ht]
    \centering
    \includegraphics[width=0.9\textwidth]{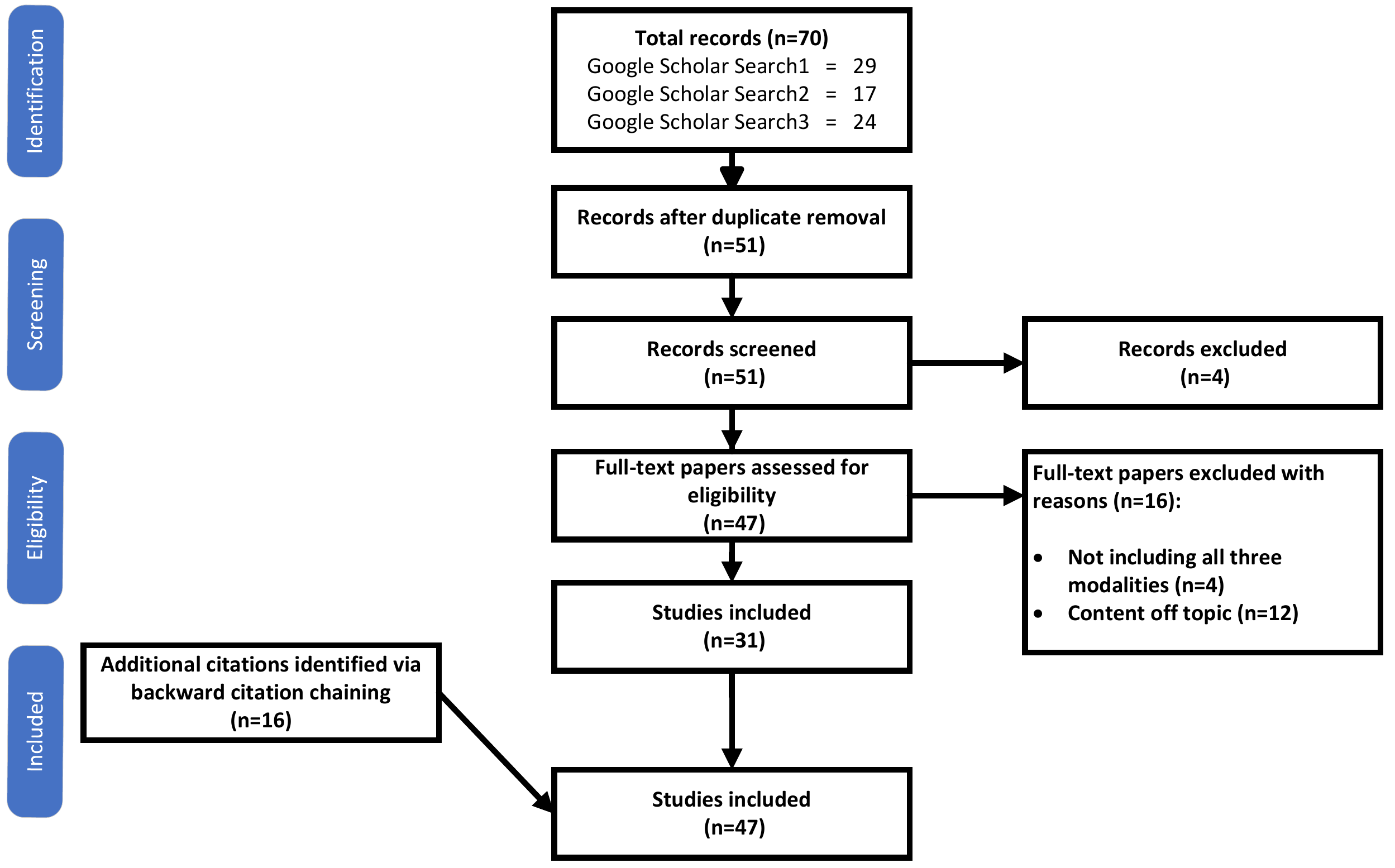}
    \caption{PRISMA flow diagram illustrating the search strategy and providing the phases of article identification and selection, which resulted in the identification of 47 papers that were deemed eligible for inclusion in the review. Prepared in accordance with Tricco AC, et al. PRISMA Extension for Scoping Reviews (PRISMA-ScR)\cite{RN137}}
    \label{fig:000 PRISMA flow diagram}
\end{figure*}
Overall, the fusion of multiple modalities is an important area of research with many potential applications in various fields. With the advancements in deep learning, it is now possible to construct more advanced systems that can perform complex tasks using fused RGB-DT data. 

\subsection*{Contribution}
\hl{
The primary aim of this survey is to provide a comprehensive and all-encompassing overview of the use of thermal cameras in combination with RGB and depth data. At the time of writing, the authors were unaware of comparable surveys specifically focusing on these technologies. While there are numerous reviews on sensor fusion, they predominantly focus on the amalgamation of LiDAR, Radar, RGB and other sensors, particularly within the realm of autonomous driving. These reviews often delve into the integration of these various sensor modalities and their specific challenges, yet they do not explore the specific tri-modal fusion of RGB, Depth (D), and Thermal (T) sensors. Our review uniquely situates itself at this intersection of sensor fusion, thereby distinguishing it from the broader landscape of sensor fusion literature.}

\hl{The primary contributions of this review paper are designed to inform  researchers working in this field by:}
\begin{itemize}
    \item \hl{Presenting a summary of various traditional and current methodologies being utilised.}
    \item \hl{Identifying available datasets for furthering this research.}
    \item \hl{Highlighting the current research trajectories and various application areas.}
\end{itemize}
\hl{
The ultimate goal is to provide a comprehensive resource that will ease the entry of interested researchers into this field while identifying trends for others.
}

As illustrated in Fig. \ref{fig:000 paper_structure}, the paper's structure begins with an introduction followed by a brief background to provide further context. Camera calibration and image registration are reviewed first since they are prerequisites for most approaches and fields of application. The discussion then shifts to modality fusion in general before examining the overlaying of thermal data onto visual data or 3D models for visual inspection or the extraction of thermal data from specific regions of interest. The use of one modality to support another in preprocessing is briefly addressed, followed by the exploration of RGB-DT applications in 3D reconstruction. Subsequently, the paper delves into manual descriptor-based methods and deep learning-based methods. Lastly, available datasets, limitations, and conclusions are presented.

\section{Methodology And Research Description}
\label{Methodology}
The systematic literature review (SLR) for this study employed the PRISMA (Preferred Reporting Items for Systematic Reviews and Meta-Analyses) methodology \cite{RN137}, which is a widely-used approach that involves a structured process for conducting a comprehensive literature search, applying eligibility criteria, extracting data, synthesising findings, and ensuring the search is reproducible with the same steps, keywords, and tags. The review began by defining the research topic of: RGB-D And Thermal Sensor Fusion. This was then followed by the definition of keywords and search tags used to search scientific databases via Google Scholar.\\
\begin{table*}[ht]
    \centering
    \caption{Literature searches on Google Scholar}
    \begin{tabularx}{\textwidth}{l>{\hsize=0.9\hsize}Xcc}
        \toprule
        \textbf{Date} & \textbf{Terms} & \textbf{Filter} & \textbf{Results} \\
        \midrule
        14.1.2023 & allintitle: thermal ( "fuse" OR "fusing" OR "object detection" OR "object-detection" OR detection) ("3D" OR "depth" OR "point cloud" OR "point clouds") & 2018 & 29 \\
        29.1.2023 & allintitle: thermal rgb-d & - & 16 (-1 duplicate) \\
        6.3.2023 & allintitle: thermal rgb "rgb-d" OR depth OR "rgb-dt" OR "rgb-d-t" & - & 6(-18 duplicates)\\
        \bottomrule
    \end{tabularx}
    \label{tab:SLR queries}
\end{table*}
\\
A comprehensive search resulted in the identification of 70 research papers related to the chosen topic. These papers were further refined by utilising exclusion criteria, such as language, repeated papers, and eliminating papers that were not relevant to the techniques under review, as depicted in the PRISMA flow diagram in Fig. \ref{fig:000 PRISMA flow diagram}.
Following the implementation of the exclusion criteria, 31 papers were reviewed in detail. Additionally, 16 more relevant documents were added after analysing the references of the initially identified papers, bringing the total number of papers reviewed to 47. Further studies that did not precisely match the three modalities, but were considered relevant to support the topic, were also included in this review. A list of the studies that have been included in the analysis can be found in Table \ref{tab:OverviewStudies}, along with additional details such as the type of sensors used, their resolution, the frequency of data acquisition, the fusion method employed, and whether or not the system is capable of real-time processing.
\begin{figure*}[ht]
    \centering
    \includegraphics[width=0.9\textwidth]{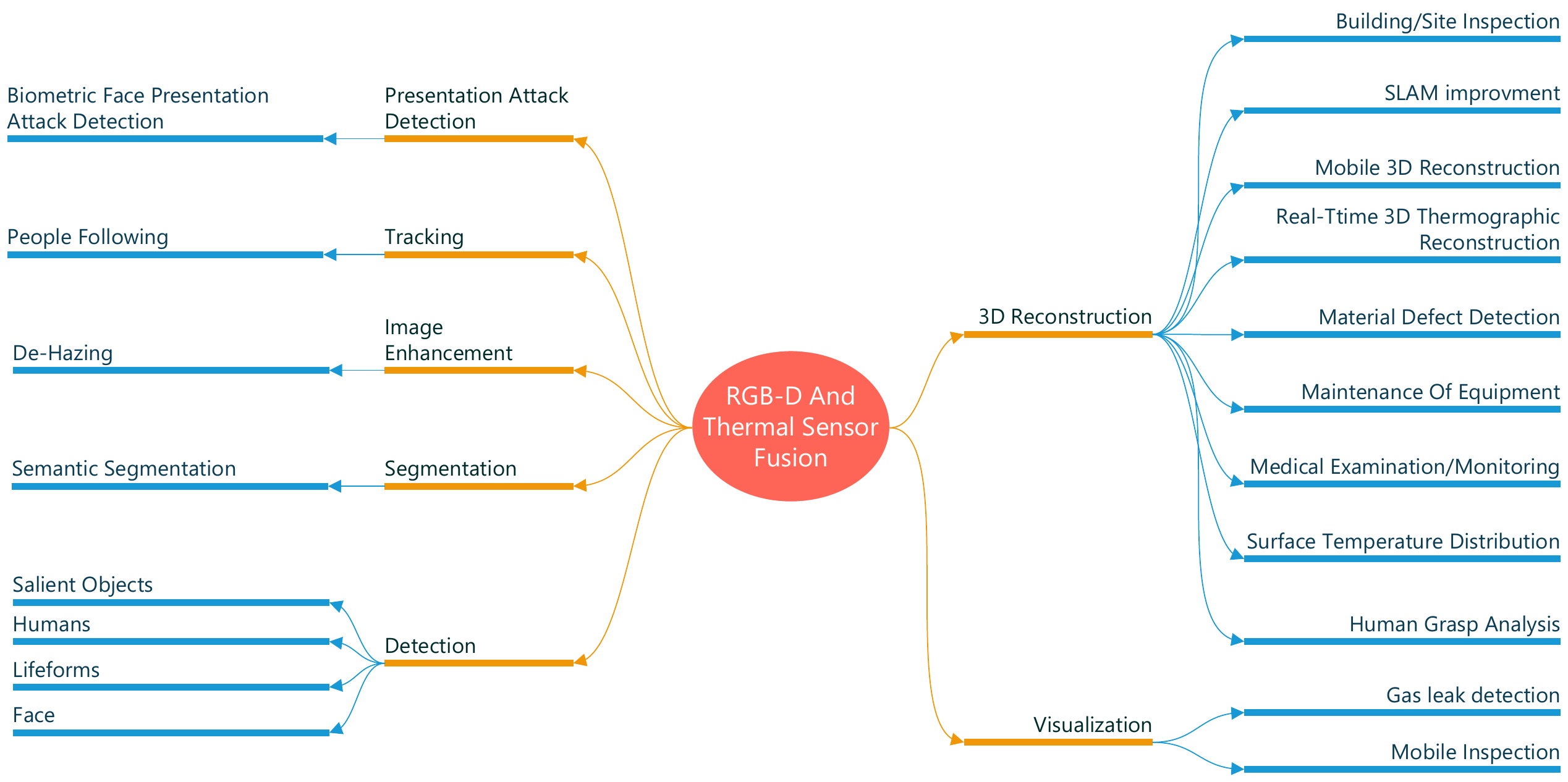}
    \caption{Overview of methods and areas of application of the reviewed documents.}
    \label{fig:000 SLR-Overview}
\end{figure*}
\subsection*{Review questions}
\noindent In this work, the aim is to answer the following review questions:
\begin{itemize}[leftmargin=*]
    \item What datasets are currently available for RGB-DT and what scenarios do they cover?
    \item What are the different methods for fusing the modalities?
    \item How are modalities weighted during fusion?
    \item What are the most suitable fusion and detection methods for real-time applications?
    \item What are the potential application areas for this technology?
    \item What are the limitations and future prospects? 
\end{itemize}

\begin{table*}[htbp]
  \centering
  \caption{Overview of reviewed studies}
  \renewcommand{\arraystretch}{0.9}
  \setlength{\tabcolsep}{4pt}
  \tiny
  \begin{tabularx}{\textwidth}{n{0.1}n{0.15}n{0.5}n{0.35}n{0.6}n{0.5}n{0.3}n{0.15}n{0.15}n{0.15}n{0.15}n{0.15}n{0.15}n{0.7}}
	\toprule
    \textbf{Ref} &\textbf{Year} &\textbf{Method} &\textbf{Datasets} &\textbf{RGB(D)} &\textbf{Thermal} &\textbf{T Res} &\textbf{T Hz} & \textbf{Type} & \textbf{Reg} & \textbf{T Use} & \textbf{Fus.} & \textbf{RT} & \textbf{ CPU / GPU} \\
    \midrule
    \cite{RN93} & 2023 & Segmentation & none & RGB Stereo & HV DS-2TD2636 & 384x288 & 50 & DL & Align & DL & E  & Y  & Unspec / GTX 2080Ti \\
    \midrule
    \cite{RN38} & 2022 & Detection general & own public & Kinect2 & FLIR A655sc & 640x480 & 50 & DL & Align & DL & MDL & N  & Unspec /Unspec \\
    \midrule
    \cite{RN42} & 2022 & Image de-hazing  & public & n/a & n/a &  &    & DL & Align & P/DL & L  & N  & Unspec / Unspec \\
    \midrule
    \cite{RN45} & 2022 & Face detection & own & Orbbec Astra & FLIR Lepton 3.5 & 160x120 & 8.7 & DL & n/a & n/a & n/a & n/a & i7-6700HQ  2.6GHz / Unspec \\
    \midrule
    \cite{RN46} & 2022 & Lifeform detection & own & epc635 & FLIR Lepton 3.5 & 160x120 & 8.7 & F  & Align & P  & E  & L  & i7-8700 / GTX 1080Ti \\
    \midrule
    \cite{RN69} & 2022 & Human detection & own & Kinect1 & Seek C. Pro  & 320x240 & 8  & DL & Align & DL & E  & L  & Unspec / Unspec \\
    \midrule
    \cite{RN72} & 2022 & 3D Reconstruction & none & Realsense D455 & FLIR Boson 320 & 320x256 & 60 & OL & Align & P/D & E  & YG & Unspec / Unspec \\
    \midrule
    \cite{RN43} & 2021 & 3D Reconstruction & none & Kinect1 & Optris PI400 & 382x288 & 80 & SW & Feature & D  & E  & N  & Unspec / Unspec \\
    \midrule
    \cite{RN44} & 2021 & Human detection & public & n/a & n/a &  &    & DL & Align & DL & EL & L  & Unspec / 2x  Titan Xp  \\
    \midrule
    \cite{RN52} & 2021 & Visualization & none & RGB Stereo & Seak & 320x240 & 15 & P  & Align & D  & OL & n/a & Unspec / Unspec \\
    \midrule
    \cite{RN91} & 2021 & 3D Reconstruction & own & DJI Zenmuse XT2  & FLIR XT2 & 640x512 & 9  & P  & Feature & D  & OL & N  & Unspec / Unspec \\
    \midrule
    \cite{RN92} & 2021 & Segmentation & public & n/a & n/a &  &    & DL & Align & DL & MF & Y  & Unspec / Unspec \\
    \midrule
    \cite{RN99} & 2021 & 3D Reconstruction & own & Realsense d415 & Optris Pi640 & 640x480 & 125 & DL & Align & D  & OL & Y  & Unspec / Unspec \\
    \midrule
    \cite{RN105} & 2021 & 3D Reconstruction & none & Realsense & FLIR A65 & 640x512 & 30 & ICP & Align & D  & OL & YG & Unspec / Unspec \\
    \midrule
    \cite{RN39} & 2020 & 3D Reconstruction & none & Photogrammetry & FLIR Zenmuse XT & 640x480 & 30 & SW & Feature & D  & OL & N  & Unspec / Unspec \\
    \midrule
    \cite{RN40} & 2020 & 3D Reconstruction & none & Photogrammetry & FLIR A65 & 640x512 & 30 & F  & Feature & D  & OL & N  & i7-10870H / RTX 2060 \\
    \midrule
    \cite{RN41} & 2020 & Tracking & none & Kinect1 & FLIR Lepton 2.5 & 80x60 & 9  & SW & Align & F  & E  & N  & Unspec / Unspec \\
    \midrule
    \cite{RN48} & 2020 & Human detection & public & n/a & n/a &  &    & DL & Align & DL & EML & L  & Unspec / Unspec \\
    \midrule
    \cite{RN49} & 2020 & 3D Reconstruction & none & Photogrammetry & FLIR E6 & 160x120 & 9  & SW & Feature & D  & OL & N  & Unspec / Unspec \\
    \midrule
    \cite{RN67} & 2020 & Human detection & none & Kinect1 & FLIR A320 & 320x240 & 9  & F  & Align & D  & L OL & Y  & Unspec / Unspec \\
    \midrule
    \cite{RN75} & 2020 & Face detection & own & n/a & n/a &  &    & DL & Align & DL & M  & n/a & Unspec / Unspec \\
    \midrule
    \cite{RN78} & 2020 & Face detection & none & Kinect1 & Optris PI450 & 382x288 & 27 & P  & Align & F  & OL & Y  & Unspec / Unspec \\
    \midrule
    \cite{RN141} & 2020 & PAD & own & RealSense SR300 & Seek C. Pro  & 320x240 & 15 & DL & Align & DL & E M & L  & Unspec / Unspec \\
    \midrule
    \cite{RN142} & 2020 & PAD & own & n/a & n/a &  &    & DL & Align & DL & E M & L  & Unspec / Unspec \\
    \midrule
    \cite{RN73} & 2019 & Face detection & none & Kinect1 & FLIR Lepton 2.5 & 80x60 & 9  & P  & Align & D  & OL & Y  & Unspec / Unspec \\
    \midrule
    \cite{RN95} & 2019 & 3D Reconstruction & none & DAVID 3D & n/a &  &    & F  & Feature & F  & OL & n/a & Unspec / Unspec \\
    \midrule
    \cite{RN119} & 2019 & PAD & own & RealSense SR300 & Seek C. Pro  & 320 × 240 & 15 & DL & Align & DL & E M & L  & i7-4800MQ / GTX 780M \\
    \midrule
    \cite{RN139} & 2019 & 3D Reconstruction & own & Kinect2 & FLIR Boson 640 & 640x512 & 30 & ICP & Feature & D  & OL & N  & Unspec / Unspec \\
    \midrule
    \cite{RN59} & 2018 & Human detection & none & Realsense R200 & FLIR Boson & n/a &    & P  & Align & DL & L  & L  & i7-9700K / RTX 2060 \\
    \midrule
    \cite{RN71} & 2018 & ROI Face detection & none & Asus Xtion & Optris PI640 & 640x480 & 32 & SW & Align & D  & OL & N  & Unspec / Titan RTX \\
    \midrule
    \cite{RN74} & 2018 & Human detection & own & Kinect2 & Optris PI640 & 640x480 & 32 & DL & Align & F  & L  & N  & Unspec / GTX 1080Ti \\
    \midrule
    \cite{RN94} & 2018 & 3D Reconstruction & none & FLIR One & FLIR One & 160x120 & 8.7 & F  & Align & D  & OL & N  & iPhone SE 1 (A9) / GT7600\\
    \midrule
    \cite{RN120} & 2018 & 3D Reconstruction & none & Kinect2 & Xenics Gobi 640  & 640x480 & 60 & ICP & Align & D  & OL & YG & Unspec / Unspec \\
    \midrule
    \cite{RN70} & 2017 & 3D Reconstruction & own & RealSense SR300 & FLIR One & 160x120 & 8.7 & P  & Align & DL & E  & N  & Unspec / GTX 680M \\
    \midrule
    \cite{RN68} & 2016 & Tracking & none & Kinect2 & FLIR A655sc & 640x480 & 50 & F  & Align & F  & MF & n/a & Unspec / Unspec \\
    \midrule
    \cite{RN89} & 2016 & Human detection & own public & Kinect1 & AXIS Q1922 & 640x480 & 30 & F  & Align & F  & MF & n/a & i7-960 / GTX 400 \\
    \midrule
    \cite{RN81} & 2015 & 3D Reconstruction & none & Kinect1 & Optris PI160 & 160x120 & 120 & ICP & Align & D  & OL & N  & i7 1.9GHz / GTX 1060 \\
    \midrule
    \cite{RN102} & 2015 & Face detection & none & Kinect2 & AXIS Q1921 & 384x288 & 30 & F  & Align & D  & MF & L  & Unspec / Unspec \\
    \midrule
    \cite{RN104} & 2015 & Visualization & none & ASUS Xtion Pro & Optris PI450 & 382x288 & 80 & F  & Feature & D  & OL & L  & Unspec / Unspec \\
    \midrule
    \cite{RN96} & 2014 & 3D Reconstruction & none & ASUS Xtion Pro & Optris PI450 & 382x288 & 80 & ICP & Feature & D  & OL & YG & Unspec / Unspec \\
    \midrule
    \cite{RN140} & 2014 & 3D Reconstruction & none & Kinect1 & Jenoptik IR-TCM & 640x480 & 60 & OL & Align & D  & OL & n/a & i7-6700K 4GHz / GTX 1080 \\
    \midrule
    \cite{RN79} & 2013 & 3D Reconstruction & none & Kinect1 & TM Miricle 307K  & 640x480 & 240 & OL & Align & D  & OL & L  & Unspec / Unspec \\
    \midrule
    \cite{RN80} & 2013 & Tracking & none & Kinect1/Hokuyo  & Heimann HTPA  & 32x31 & 9.1 & F  & None & F  & L  & Y  & i7-6700K 4GHz / GTX 1080 \\
    \midrule
    \cite{RN88} & 2013 & Re-identification & own & Kinect1 & AXIS Q1922 & 640x480 & 30 & F  & Align & F  & MF & L  & Unspec / Titan Xp \\
    \midrule
    \cite{RN97} & 2012 & 3D Reconstruction & none & Riegl VZ-400 laser & Optris PI160 & 160x120 & 120 & P  & Align & D  & OL & n/a & Unspec / Unspec \\
    \midrule
    \cite{RN98} & 2012 & 3D Reconstruction & none & RGB FLIR E60 (SfM) & FLIR E60  & 320x240 & 60 & F  & Feature & D  & OL & n/a & Unspec / Unspec \\
    \midrule
    \cite{RN101} & 2011 & 3D Reconstruction & none & Kinect1 & TC384 & 384×288 & 50 & OL & Align & D  & OL & n/a & Unspec / Unspec \\
    \bottomrule
 \end{tabularx}%
 \vspace{1ex}
  \par\noindent
  \footnotesize\raggedright
  \textcolor{black}{
  "T Use": Thermal data application; Display(D), Post-processing or Process/Algorithm(P), Feature(F), Deep Learning(DL)\\
            "Fus.": How the Thermal data was fused with the other modalities; Late(L), Middle(M), Early(E), Overlay/Align(OL), Feature(F)\\
            "RT": Inference speed real-time (>=30FPS); Yes(Y), Yes with GPU(YG), Likely but no data provided(L), No(N)\\
            "Reg": Registration process: Image alignment (Align), Feature matching (Feature)\\
            "Unspec": Unspecified}
 \label{tab:OverviewStudies}%
\end{table*}%

\section{Background}

The initial research papers that concentrated on fusing RGB, Depth, and Thermal data (RGB-DT) using RGB-D cameras emerged in 2011. Early works in this field investigated medical scans \cite{RN101}, while later in 2013, research expanded to include 3D thermal mapping of building interiors \cite{RN79}, sensor fusion for people tracking \cite{RN80}, and tri-modal person re-identification \cite{RN88}.

Some earlier works proposed systems using different technologies, such as a terrestrial laser scanner and thermal infrared camera \cite{RN97}, or a Structure from Motion (SfM) or MultiView Stereo (MVS) pipeline to generate a dense, coloured point cloud with optional thermal data overlay \cite{RN98}. 

Over the last decade then, the fusion of multiple modalities has been increasingly researched as combining different modalities, such as RGB-DT data, has been recognised to provide a richer and more comprehensive representation of the environment or scene. This has resulted in achieving a more accurate and robust performance in a wide range of applications, including building mapping\cite{RN79}, person re-identification\cite{RN88}, 3D salient object detection\cite{RN38}, autonomous driving\cite{RN42,RN112}, activity recognition\cite{RN74}, robotics\cite{RN38}, surveillance\cite{RN44}, 3D reconstruction\cite{RN72,RN96}, defect detection\cite{RN95}, gas leak detection \cite{RN104} and many others. For example, in robotics, the combination of RGB-DT data can enable robots to perceive and navigate through complex environments with greater accuracy and efficiency \cite{RN38} or to interact with humans better by interpreting their emotions\cite{RN75} and activities\cite{RN74}. In surveillance, the fusion of RGB-DT data has been shown to improve the detection and recognition of objects, people, and activities in a monitored area \cite{RN68} under difficult light conditions, and in autonomous driving, the fusion of multi-modal data can provide a more comprehensive understanding of the surrounding environment, enabling safer and more reliable driving\cite{RN112}\cite{RN42}. The temperature characteristics of maize under water stress, which serves as an example of multi-modal sensing in agriculture, has been investigated \cite{RN43}, which has the benefit of developing more efficient and sustainable agricultural practices. In the field of industrial maintenance, an Augmented Reality(AR) system, that visualises components and their temperature in real-time has been proposed \cite{RN99}, helping to identify faults and problems. It is also worth noting that in some cases, a single modality can indirectly improve the quality of another modality. This has been demonstrated in \cite{RN42} by using the thermal data, and the extracted monodepth\cite{RN117} data from the thermal data, in an algorithm used to dehaze the RGB image so that it can be used for object detection further down in the processing pipeline. Fig. \ref{fig:000 SLR-Overview} offers a summary of the methods and application areas covered in the reviewed documents. This figure highlights the wide range of approaches and techniques employed in the research papers, as well as the diverse fields where these methods have been implemented.

While early works employed traditional computer vision techniques, the field has evolved alongside advancements in deep learning. Although the fusion of modalities has been shown to outperform single-modality systems, only a limited number of researchers have tackled the topic of heterogeneous sensor fusion involving stereo vision or depth cameras with thermal cameras. This is despite the growing need to meet evolving requirements and develop more robust decision-making systems by integrating features from various sensors. The potential for improved performance in a range of applications highlights the importance of continuing to explore and develop these multi-modal fusion approaches. Figure \ref{fig:000 NumberOfStudiesRGBDT} shows the evolving trends for RGB-D, RGB-T and RGB-DT research by depicting the number of studies published over the past 10 years. 
\section{Camera Calibration And Registration}
\begin{figure}[ht]
    \centering
    \includegraphics[width=0.48\textwidth]{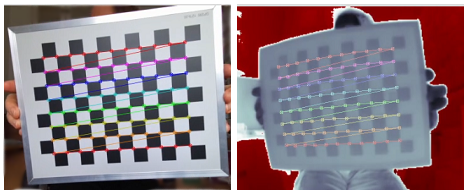}
    \caption{\hl{Stereo calibration of RGB and Thermal cameras. The left image shows a heated bi-material calibration checkerboard captured by an RGB camera, while the right image presents the same board as seen by a thermal camera. The overlaid lines illustrate the pattern recognition process of the stereo calibration.}}
    \label{fig:000 StereoCalibration}
\end{figure}

For successful multimodal environmental sensing using RGB, depth, and thermal data, it is crucial to acquire the data from these modalities in a properly aligned manner. This can pose a challenge since the sensors used for each modality may have varying fields of view (FOV), resolutions, and sensing capabilities. To facilitate data fusion, the system must be calibrated by determining the intrinsic (pin-hole camera model parameter matrix) and extrinsic (estimation of the relative sensor poses) parameters of each camera, which can then be used to align the data. This calibration, based on the pinhole camera model, has been simplified by using a stereo calibration process\cite{RN100}, which can be applied using these and similar modalities. This method has been implemented in numerous studies\hl{ }in different ways. Fig. \ref{fig:000 StereoCalibration} shows the pattern matching using stereo calibration.

\subsection{Calibration Boards}
The most popular approach for the geometric calibration of thermal cameras used to be a printed chessboard heated by a ﬂood lamp which was comparatively inaccurate and difﬁcult to execute\cite{RN109} as the temperature difference was fading quickly and the pattern was blurry. To address this a novel geometric mask with high thermal contrast that does not require a flood lamp has been proposed \cite{RN109} as an alternative calibration pattern. This approach involves cutting a mask out of a thin material and holding it in front of a backdrop with a different level of thermal radiance. Building on this idea, various constructions have been developed in recent years, all based on the same principle.
\begin{figure}[ht]
    \centering
    \includegraphics[width=0.48\textwidth]{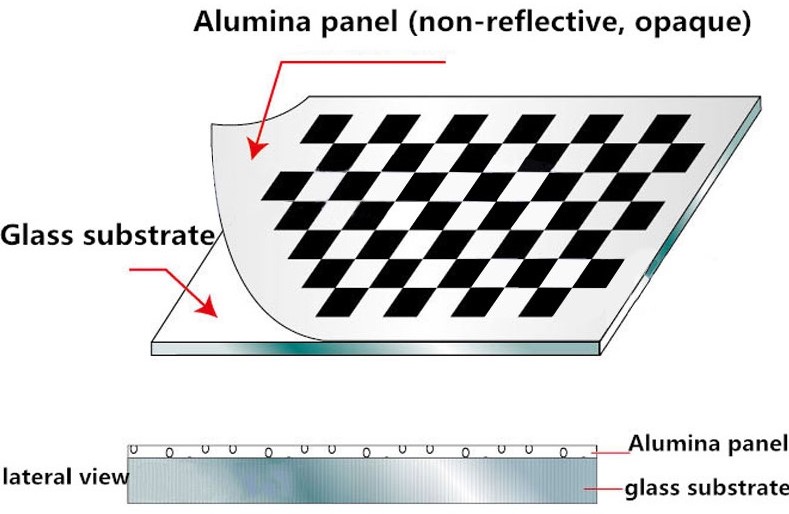}
    \caption{RGB-DT calibration board made of a glass substrate and Alumina panel.\cite{RN38}}
    \label{fig:RN38 F5 calibration}
\end{figure}

The multi-material calibration boards, which are essential for cross-calibrating thermal and visual modalities, with their distinct geometric patterns visible in all calibrated modalities, are used in the calibration process \cite{RN100}. A checkerboard with 12 × 9 (30 mm for every square grid) with the pattern printed onto an alumina plate has been used \cite{RN38} which is then mounted on a glass substrate as illustrated in Fig. \ref{fig:RN38 F5 calibration}. The board is heated from the back, while the white reflects the heat, the black conducts it to produce the pattern in the thermal modality. These boards are commercially available. The authors in \cite{RN106} constructed a board where the calibration pattern comprises a line-based grid with regularly sized square patterns. The pattern consists of thin copper lines milled onto a printed circuit board (PCB) with a width of 2 mm and a spacing of 40 mm, and it has six/seven intersections along the shorter/longer axis. Compared to conventional calibration patterns, the line-grid pattern is more robust in maintaining high contrast in thermal images due to the good conductivity of the copper lines, which ensures a uniform thermal distribution. Additionally, the proposed pattern has the same geometric relations as the conventional chessboard pattern, allowing for the use of existing algorithms for camera calibration.

\begin{figure}[ht]
    \centering
    \includegraphics[width=0.48\textwidth]{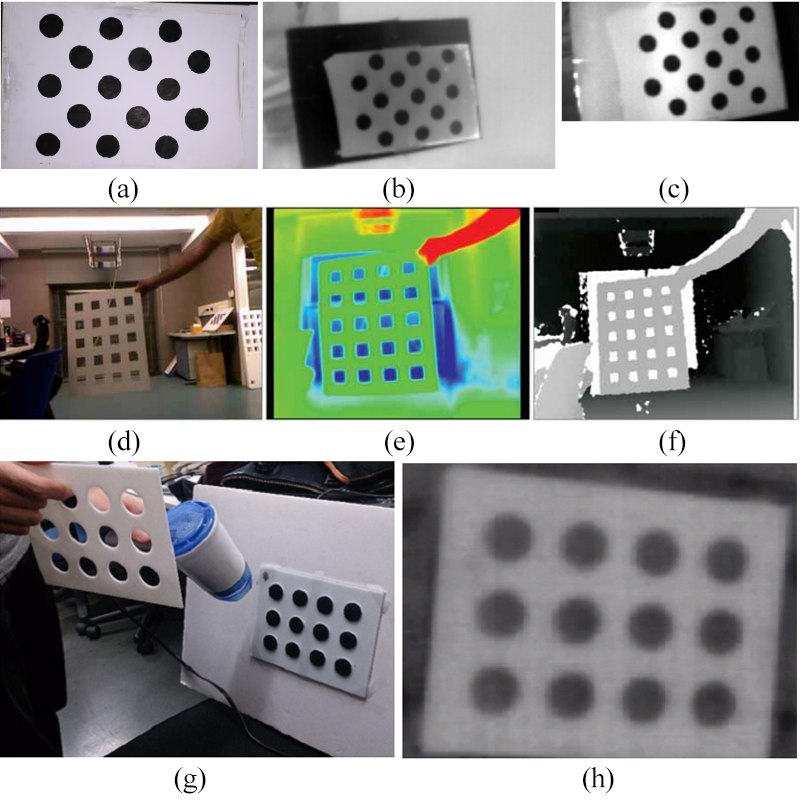}
    \caption{Calibration boards used in studies \cite{RN46}, \cite{RN101} and \cite{RN81} from top to bottom respectively: \subcaption{a}RGB \subcaption{b}Thermal \subcaption{c}Depth \subcaption{d}RGB \subcaption{e}Thermal \subcaption{f}Depth \subcaption{g}Heating of lower plate \subcaption{h}Thermal}
    \label{fig:000 calibrationBoards}
\end{figure}

However, calibration boards can be constructed simpler as demonstrated in \cite{RN49} where an 11 × 11 checkerboard pattern made of cardboard paper and highly reﬂective metal squares was used. Alternatively, \cite{RN89} constructed the calibration board using an A3-sized 10mm polystyrene foam board as a backdrop and a board of the same size with cut-out squares as the checkerboard. This is similar to \cite{RN101} where a solid board was used that had rectangular holes cut out, as shown in Fig. \ref{fig:000 calibrationBoards}(d), whereas \cite{RN52} used fabric for the black pattern. In addition to using squares, circles can also be used, as demonstrated in \cite{RN46} where the authors utilised a mask made of 3mm thin Depron\textsuperscript{\textregistered} material with an asymmetric circle pattern, as shown in Fig. \ref{fig:000 calibrationBoards}(a), while \cite{RN81} proposed using 3D printed boards in their study, as shown in Fig. \ref{fig:000 calibrationBoards}(g).

A distinct approach was adopted in \cite{RN78} and \cite{RN41}, where resistors were placed onto the calibration board and heated up electrically, enabling a prolonged calibration process. Similarly, \cite{RN67} and \cite{RN59} employed incandescent light bulbs embedded at every other corner of the grid to emit heat.
In \cite{RN121}, a method was proposed for calibrating a UV camera with RGB-D and thermal cameras using a rectangular aluminium plate with evenly distributed circle holes. A heater strip is placed behind the plate to create sufficient contrast for the thermal camera. To ensure that all cameras can be calibrated together, a black box is used, which absorbs most of the light while allowing light to pass through the holes on the aluminium board. The white paper covering the board reflects visible and UV light, which can be detected by the RGB and UV cameras. Once the calibration tool's features are detected, the centre of each circle is marked, and OpenCV's \cite{opencv_library} camera calibration function is used to obtain intrinsic and transformation matrices for each camera coordinate system. The proposed method allows for the accurate calibration of multiple cameras, including a UV camera, which can be beneficial in various applications. \cite{RN72} used this approach to calibrate RGB-D and thermal modalities.

The combination of thermal images and colour images typically involves the use of methods that require complex calculations. However, in\cite{ben2016epipolar}, a 2-point approach was proposed that outperformed commonly used 8-point and 7-point approaches for equalising the epipolar geometries of different images. The study proposes a method for effectively combining images by determining two points on the epipolar plane. This technique was also employed in \cite{RN69} for the calibration process. In order to find calibration points in both thermal and optical data, the method used in the study involves several operations. Firstly, the Canny Edge detection method is applied to the thermal image to determine the calibration points. Next, in the optical image, the Hough circle finding method is used to locate the circles containing the calibration points, and the centres of these circles are determined as calibration points. \hl{It is important to note that the calibration mechanism design consists of two black circle drawings on a white background with incandescent bulbs at the centre of these circles. This setup allows for the creation of distinguishable common points in both the thermal and optical data, which are essential for the calibration process.} Following this, line segments are extracted and plotted on both thermal and optical data. The lengths of these segments are determined by the Euclidean distance, and the slopes of the lines between the points are calculated using the slope formula and are stored for the combining process. The rotation of the thermal image is based on the difference in the calculated slopes of the lines, followed by resizing the thermal image with respect to the line length ratio. The midpoints and distances between them are obtained from thermal and optical images to achieve precise alignment in the same plane. This allows the determination of the position of the thermal image relative to the optical image\cite{RN69}.

In some RGB-D sensors, like the Microsoft Kinect series, the depth stream originates from a time-of-flight camera that also generates an additional IR stream from amplitude information. Since both streams originate from the same sensor, it is referred to as the Depth/IR sensor. The IR stream can be utilised for calibration purposes, eliminating the need for any 3D elements on a board and can provide supplemental data that may be beneficial in applications for object detection or tracking in low-light conditions. This IR stream senses the 850nm (NWIR) spectral band and does not contain any thermal data. It is important to clarify that this stream should not be mistaken for the stream from a thermal camera, which is based on wavelengths of roughly $8-14\mu$m (LWIR). An overview of the spectral range is given in Fig. \ref{fig:000 spectral_range}.
\subsection{Registration}
RGB-D cameras, including models like Microsoft Kinect (V1, V2, and Azure) and Intel RealSense (D415, D435, etc.), are engineered to simultaneously capture both visual and depth modalities. As a result, they inherently register and output both data types. To align the thermal data, the stereo calibration process can be used to register it against the visual data.

In earlier works, before calibration using geometric patterns was applied, researchers used the Hough Parameter Space to register modalities as demonstrated in \cite{RN100}. This process involved detecting edges with the Canny edge detector, resulting in binary edge images. These images were then processed by the Hough transform, which extracted all linear image segments. The rotation and translation differences could be calculated using line correspondence analysis \cite{RN108}.
Nonetheless, considering the two modalities as a stereo pair and employing stereo calibration techniques simplifies this process. The algorithm \cite{RN100} has since been conveniently integrated into various tools such as OpenCV \cite{opencv_library}, Matlab \cite{MATLAB}, and other tools and frameworks, facilitating the acquisition of the translation vector, rotation matrix, and distortion coefficients.

\hl{The calibration of multi-camera systems, each characterised by a unique field of view (FOV), can be a challenging task, particularly when it involves a variety of modalities and resolutions. RGB and RGB-D cameras typically offer higher resolutions and distinct FOVs compared to thermal cameras. For accurate sensor fusion, optimising the overlap between the RGB, depth, and thermal modalities is crucial. In RGB-D cameras, the RGB component is usually internally pre-adjusted to match the overlapping FOV of the depth data.
In sensor fusion processes that are designed for subsequent analysis and necessitate overlapping data from all modalities for real-time processing, it is necessary to modify the RGB-D data through cropping or clipping to match the resolution and FOV of the thermal camera. This requires careful consideration of the FOV of each camera during system design to ensure maximum overlap. When aligning a lower-resolution image with a smaller FOV to a higher-resolution image with a larger FOV, a homography is typically used to transform the lower-resolution image to align with the corresponding part of the higher-resolution image. This approach, which allows for the incorporation of additional information from the lower-resolution image while preserving the high-resolution data, is employed in the study }\cite{RN81}\hl{. In this study, the authors effectively align and fuse data from sensors with different FOVs and resolutions. They further address the challenges of occlusions and significant differences in the FOVs of the cameras, demonstrating the versatility and robustness of this approach in handling complex sensor fusion scenarios. 
In offline processing scenarios, more intricate techniques can be employed for alignment. For instance, the paper } \cite{RN91}\hl{ utilises a combination of Scale-Invariant Feature Transform (SIFT) for keypoints computation and matching, Random Sample Consensus (RANSAC) for eliminating geometrically inconsistent matches, and Bundle Block Adjustment (BBA) for optimising camera parameters and producing an initial 3D structure of aligned images. Although these methods are computationally demanding, they offer superior accuracy and robustness, making them ideal for applications where precision is crucial.}

\hl{While the stereo calibration approach is effective, it encounters a significant challenge from the different FOVs of the modalities, which can result in a parallax effect that varies at different depths. This phenomenon is due to the difference in viewing angles between the cameras, causing objects at various depths to appear at different positions in the different cameras. As a result, using a single homography, a transformation that maps points in one image to corresponding points in another image, only functions effectively on a specific plane. This variation in perspective leads to misalignment in the fused data.} 

\hl{One approach to overcoming this problem is presented in} \cite{RN89} \hl{. Firstly, a thermal-visible calibration device is used to establish the correspondences between the points extracted from the thermal and RGB modalities. Using a Microsoft Kinect camera, the depth sensor is already factory registered to the RGB camera; therefore, registration is focused only on the RGB to thermal data.  Registration is performed using a weighted sum of multiple homographies. Multiple views of the calibration device scattered throughout the exploratory scene were used to generate homographies relating RGB and thermal modalities. Each homography is calculated using a RANSAC-based method, taking into account the approximate distance to the view of the calibration device represented by the homography. This strategy effectively compensates for parallax at different depths. The rationale behind the approach is that registration based on each homography is only accurate for points on the plane that are spanned by the particular view of the calibration device. Therefore, to register an arbitrary point in the scene, the 8 closest homographies are weighted and then summed up. It was observed that registration accuracy is primarily dependent on 3 factors: the distance in space to the nearest homography, the synchronisation of RGB and thermal cameras, as well as the accuracy of the depth estimate.}

In photogrammetry-based 3D reconstruction, as demonstrated in \cite{RN91}, the registration process depends on the identification and matching of keypoints, which is followed by Bundle Block Adjustment (BBA) \cite{kraus2007photogrammetry}. Keypoint computation involves the detection and description of features using the SIFT \cite{lowe2004distinctive} algorithm. Keypoints are unique locations in the image that correspond to the same real-world object across different images. The matching step entails finding matching keypoints across overlapping images. Subsequently, BBA is used to optimise the camera parameters, both internal and external, for each image, ensuring accurate calculation of ray paths inside and outside the camera for precise 3D reconstruction. These keypoint computations, matching, and BBA algorithms have been extensively studied and integrated into various software packages and frameworks for photogrammetric applications.

\paragraph*{Automatic registration}
A different approach was taken by the authors in \cite{RN43} by extracting edge images. To register, feature points were detected and matched. Common feature descriptors used for image registration include SIFT \cite{lowe2004distinctive}, SURF \cite{bay2006surf}, and BRISK \cite{leutenegger2011brisk}. However, these methods often involve the use of a Gaussian filter, which can cause the loss of image details. To address this issue, \cite{alcantarilla2012kaze} proposed a new feature descriptor called KAZE, which can detect image features in nonlinear scale spaces and obtain more feature points. The KAZE feature descriptor was utilised to register thermal and colour images of maize. The KAZE features and key points were detected from extracted edge images, and their descriptors were built. Feature points were then matched using the nearest neighbour distance ratio strategy, with outliers removed using the M-estimator Sample Consensus (MSC) algorithm, a variant of the RANSAC algorithm. This approach is akin to \cite{RN95}, which is elaborated in more detail in section \ref{subsub: Fusion Feature}. The study proposed a feature-based registration method for aligning thermal and RGB-D images using the Shape Constrained SIFT Descriptor (SCSIFT).

A similar auto registration approach was taken in \cite{RN104}, Edge-Based Mutual Information (EMI). However, they encountered issues when utilising the thermal images because of the Automatic Gain Control (AGC) employed in the thermal video stream. This AGC results in a variable colour range, as depicted in Fig. \ref{fig:000 VDT-2048 example}, which shows an example from the VDT-2048 dataset. Their proposed method combines mutual information (MI), edge detection, and image separation to achieve image registration with the following steps:

\noindent
\textbf{Image filtering:} The input images are first filtered using a Gaussian filter to reduce noise. This is done with a 9x9 kernel size and a standard deviation ($\sigma$) of 1.85. 

\noindent
\textbf{Edge detection:} A Canny edge detector is applied to both filtered images to generate edge images. 

\noindent
\textbf{Region separation:} After obtaining the edge images, region separation is performed. The primary goal of this step is to constrain the mutual information (MI) optimisation functions to focus on grey values that are in the vicinity of edges. This approach helps to ensure that the MI optimisation process is more accurate and robust, as it considers only the most relevant information in the image.

When software tools are used for the 3D reconstruction, as in \cite{RN39,RN49}, the registration algorithms are applied by those software packages and are mainly based on a combination of feature detection, feature matching, and bundle block adjustment:

\noindent
\textbf{Feature detection:} Identifies keypoints or features in each image. These features are typically distinct and easily recognisable patterns, such as corners, edges, or textures. The software employs SIFT (Scale-Invariant Feature Transform) or similar algorithms to extract features from the images.

\noindent
\textbf{Feature matching:} After detecting features in each image, match corresponding features across multiple overlapping images. The software uses a matching algorithm, such as approximate nearest neighbour matching, to find the best matches between the features detected in different images.

\noindent
\textbf{Bundle block adjustment:} Once the matching features have been identified, employ a bundle block adjustment technique to optimise the camera positions and orientations, as well as the 3D coordinates of the keypoints. This process involves minimising the reprojection error, which measures the discrepancy between the observed image coordinates and the projected coordinates of the keypoints in 3D space. Bundle block adjustment refines the initial estimates of camera parameters and 3D points to improve the overall accuracy of the reconstructed scene.

By combining these techniques, the software registers the images, ultimately creating a consistent and accurate 3D representation of the surveyed area.

\section{Thermal Data Visualisation}
The Automatic Gain Control (AGC) technique is a histogram-based processing method that transforms raw data formats into 8-bit image data. However, this processing results in data compression, leading to a significant loss of information. In the case of 16-bit data, with a possible value range of 0 to 65,535, the resulting image is represented with values in the 0 to 255 interval, further decreasing detail. To address this issue, AGC algorithms are designed to enhance image contrast and brightness, thereby emphasising the contextual details of the scene \cite{RN143}.

Most LWIR cameras produce a grayscale or colour-range image stream with 8-bit per pixel. They typically use an AGC algorithm to generate the 8-bit image with high contrast. The 8-bit data represents gain-controlled values that depend on the temperature of objects in the scene and are more appropriate for human vision. However, the 8-bit representation results in a lower thermal resolution and the algorithm causes colour changes based on minimum and maximum measurements.
\section{How And What To Fuse}
In multimodal sensor fusion, deciding how and what to fuse depends on the specific application, the data modalities involved, and the desired outcome. The fusion of features or decisions can be achieved in many ways, such as concatenating feature vectors, averaging or weighted averages of data or decisions, weighted voting schemes to combine decisions, or applying machine learning techniques such as neural networks, decision trees, or support vector machines. However, the fast and massive data collection capabilities of the sensors and the representation of the obtained large data in the memory, possibly with different data types, are one of the challenges of real-time sensor fusion\cite{RN69}.

\hl{Alongside the fusion of different modalities, another important aspect to consider is the methodology of sensor fusion implementation. Two primary approaches dominate this field: model-based and data-driven techniques. Model-based methods, as explored in study }\cite{RN145}\hl{, utilise pre-established models to interpret and integrate sensor data. These methods often exhibit robustness and interpretability, but their effectiveness can be constrained by the accuracy of the models they employ. In contrast, data-driven techniques, as outlined in the research }\cite{RN146}\hl{, learn to merge sensor data directly from the data itself, typically employing machine learning techniques. These methods can potentially achieve superior performance, but they may require substantial data quantities and may be less interpretable.}
\subsection{Fusion Stages}
\label{subsec:FusionStages}
Features of multiple modalities can be fused at different points in a process, and these fusion points are generally categorised into three levels: Data level, Feature level, and Decision level. These levels can also be referred to as low, mid, and high or early, middle, and late fusion. Each level of fusion has its advantages and disadvantages so it is essential to consider the specific context when selecting the fusion point.
The three levels can be categorised as:

\begin{enumerate}
    \item Data level(early) fusion:
At the data level, the fusion of different modalities involves combining raw data from all modalities to create an integrated dataset, often by concatenating or averaging. This approach is useful when the raw data from different modalities are directly comparable and compatible. For RGB-DT data, often multi-channel images are created by blending and combining the data, primarily for deep learning purposes\cite{RN69,RN44,RN48}.

    \item Feature level(middle) fusion:
In this approach, features are extracted separately from each modality and then combined before being fed into a classifier or a learning algorithm. Feature-level fusion can involve concatenating the feature vectors or using other methods to merge the extracted features. This method often results in a more compact and informative representation of the data, as the features from each modality are combined after being extracted, retaining information specific to each modality. In manually crafted feature-based approaches, this is a common approach while in deep learning, this method usually enhances accuracy but has higher computational requirements. There are many variations of middle fusion depending on the processing pipeline. \hl{The authors of} \cite{RN68} \hl{propose an algorithm optimised for human tracking based on an enhanced Bhattacharyya coefficient, and in }\cite{RN89} \hl{features are fused for body segmentation using stacked learning and Random Forest while in }\cite{RN102} \hl{features are combined by applying landmark-based energy filters for pain level recognition.}

    \item Decision level(late) fusion:
At this stage, each modality is processed separately, with features extracted and then classified or analysed independently. The results or decisions from each modality are then combined to produce a final decision or output. This approach is suitable when the modalities are diverse and difficult to compare directly, or when separate classifiers have been optimised for each modality. \hl{Decision-level fusion can involve using majority voting}\cite{RN40,RN88}\hl{, weighted voting}\cite{RN46}\hl{, or other decision-fusion techniques like a Support Vector Machine(SVM)}\cite{RN74}.
\end{enumerate}

The performance of a fusion method is highly dependent on the sensing modalities, data, and network architectures being used. This rough categorisation also holds true when applying Deep Neural Networks(DNN), which is discussed in more detail in section \ref{subsec:DeepLearningMethods} and in section \ref{sub:Feature-Based Methods} for the feature-based approach. The fusion of the modalities however is not limited to a single stage but can be applied at multiple stages in a processing pipeline.
Besides the listed fusion methods, it is also worth mentioning that the direct fusion of multiple modalities is not the only way to enhance the quality of data. A single modality can also indirectly enhance the quality of another modality. In \cite{RN42} for example, the authors used the thermal data, together with the monodepth\cite{RN117} data extracted from it, to improve the quality of RGB images by applying a dehazing algorithm. 

\subsection{Fusion Methodologies}
\label{subsec:FusionMethodologies}
\hl{Sensor data fusion methodologies can be broadly categorised into two main approaches: model-based and data-driven. }

\begin{itemize}
    \item \textbf{Model-Based Approaches} These methods rely on predefined models to interpret and combine sensor data. They are often robust and interpretable but may be limited by the accuracy of the models they use. Some common techniques under this category include:
    \begin{itemize}
    \item \textbf{Kalman Filters} These are utilised in linear systems characterised by Gaussian noise, offering optimal performance in terms of minimising the mean squared error. As demonstrated in the study by the authors of \cite{RN80}, a Kalman filter, when combined with a probabilistic model of a leg shape, can ensure robust tracking in scenarios such as person-following.
    
    \item \textbf{Particle Filters} These are used for non-linear and non-Gaussian systems. They are more flexible than Kalman filters but require more computational resources. In the context of person tracking, the authors of \cite{RN68} employed a simple particle filter approach, which estimates the target's probability distribution using a set of weighted particles while study \cite{RN41} presents an adaptive human tracking method using. The method incorporates adaptive weighting based on velocity and head position, allowing it to handle fast motion, partial occultation, and scale variation. The fusion of depth and thermal data enhances the robustness and accuracy of the tracking process, as demonstrated in various challenging scenarios.
    
    \item \textbf{Bayesian Networks} These models are utilised in probabilistic modelling to represent the probabilistic connections between a group of variables. They are particularly beneficial when the relationships between the sensors are either known or can be learned. In the domain of Presentation Attack Detection (PAD), the authors of study \cite{RN47} employed Bayesian Networks to differentiate between a genuine face and a fraudulent attack. Their approach involved designing an attack detector module based on Bayesian principles, with the decision boundary set at a log-likelihood ratio of attack to bona fide equal to 0. This design choice ensures that the classifier operates independently and maximises the confidence score in its classification.
    \end{itemize}
    
    \item \textbf{Data-Driven Approaches} These methods learn to combine sensor data directly from the data itself, often using machine learning techniques. They can achieve higher performance but require large amounts of data and can be less interpretable. Some common techniques under this category include:
    \begin{itemize}
    \item \textbf{Support Vector Machines (SVMs)} These are powerful supervised learning models that perform well in high-dimensional spaces and can be customised with different Kernel functions for the decision function. However, their effectiveness can be surpassed by more complex models such as CNNs in certain contexts, as shown in \cite{RN58} for fall detection systems. In the context of activity recognition and emotion classification, SVMs have demonstrated promising results when combined with various types of features. For example, the authors of \cite{RN74} utilised an SVM model trained with depth and skeleton features in conjunction with thermal sensor data to enhance activity recognition accuracy. Similarly, \cite{RN75} used an SVM model with both gait Power Spectral Density (PSD) and thermal features, achieving an offline testing accuracy of 70\% in emotion classification while the authors of \cite{RN89} compared their human body segmentation, based on Random Forest, with one using HOG + SVM. Despite the HOG + SVM approach being trained on larger, varied datasets, the study's proposed method significantly outperformed it. Further, the authors of \cite{RN142}, in the context of Presentation Attach Detection (PAD), noted that the SVM baseline generally performed worse than the other approaches, suggesting that the local, pixel-wise classification approach may not be as effective as the more holistic view provided by CNN models in their evaluation.

    \item \textbf{Decision Trees} These flowchart-like structures are used for decision-making, where each internal node signifies a test on an attribute, each branch represents the outcome of the test, and each leaf node holds a class label. They are valued for their simplicity and interpretability. For instance, the authors of \cite{RN146} utilised a decision tree-based algorithm in a novel data association approach. This method used polar rays to find correspondences between trifocal camera objects and fused hypothesis, or super-sensor objects. The decision tree gradually eliminated unwanted associations by considering object characteristics such as area, visible façade, dimension ratio, and relative position in different coordinate systems.
    
    \item \textbf{Random Forest} This ensemble learning method constructs multiple decision trees during training and outputs the class that is the mode of the classes of the individual trees. A practical application of this technique is demonstrated in \cite{RN40}, where a Random Forest was used to predict the conditional probabilities of different class labels based on point descriptors. The Random Forest, an ensemble of decision trees, was trained on randomly sampled subsets of training data. This approach resulted in decorrelated trees that enhanced the generalisation and robustness of the classification. The final point label was determined by majority voting across all decision trees in the Random Forest.

    \item \textbf{Neural Networks and Deep Learning Models} Neural networks excel at discerning complex patterns within high-dimensional data, such as images. Deep learning, a subset of neural networks, utilises multiple hidden layers to automatically learn and extract features from raw data, proving highly effective for RGB-D and Thermal sensor fusion tasks. A more comprehensive discussion on this topic can be found in section \ref{subsec:DeepLearningMethods}.
    \end{itemize}
\end{itemize}

\hl{It's important to note that these categories are not mutually exclusive, and sensor fusion systems may use a combination of these approaches. The choice of methodology, similar to the selection of fusion methods, depends on the specific requirements of the task, the available data, and the computational resources.}
\section{ROI \& Overlay}
In some applications, thermal data serves as supplementary information for analysis purposes, such as site \cite{RN49,RN91} and building \cite{RN39,RN40,RN96} inspections, medical examinations \cite{RN67}, or human thermal comfort assessments \cite{RN73}. Large-area inspections for sites and buildings are generally not performed in real-time or with RGB-D sensors. Instead, photogrammetry \cite{kraus2007photogrammetry} is employed, either with custom-built processing pipelines as in \cite{RN91} or established tools like Pix4Dmapper, 3DF Zephyr, Context Capture, PhotoScan, and others as in \cite{RN39,RN49}, to generate point clouds offline. By aligning thermal images, the point clouds are enriched with thermal data for offline analysis.
In contrast, \cite{RN94} used mobile devices and proposed image-based modelling (IBM), a passive mapping technique that uses image datasets with multiple fields of view (FOV) to reconstruct 3D models. This study employed a low-cost thermal camera and two smartphones to capture visible and thermal images. The work established that the proposed method is cost-effective and achieves a temperature precision of 2°C in the 3D thermal models, albeit at a slower pace.
Since these approaches are not the primary focus of this study, they are not pursued further but are mentioned for completion as they also represent a type of fusion of these modalities. However, the modalities are not fused to enhance a process but merely for post-analysis. 

\begin{figure*}[ht]
    \centering
    \includegraphics[width=0.9\textwidth]{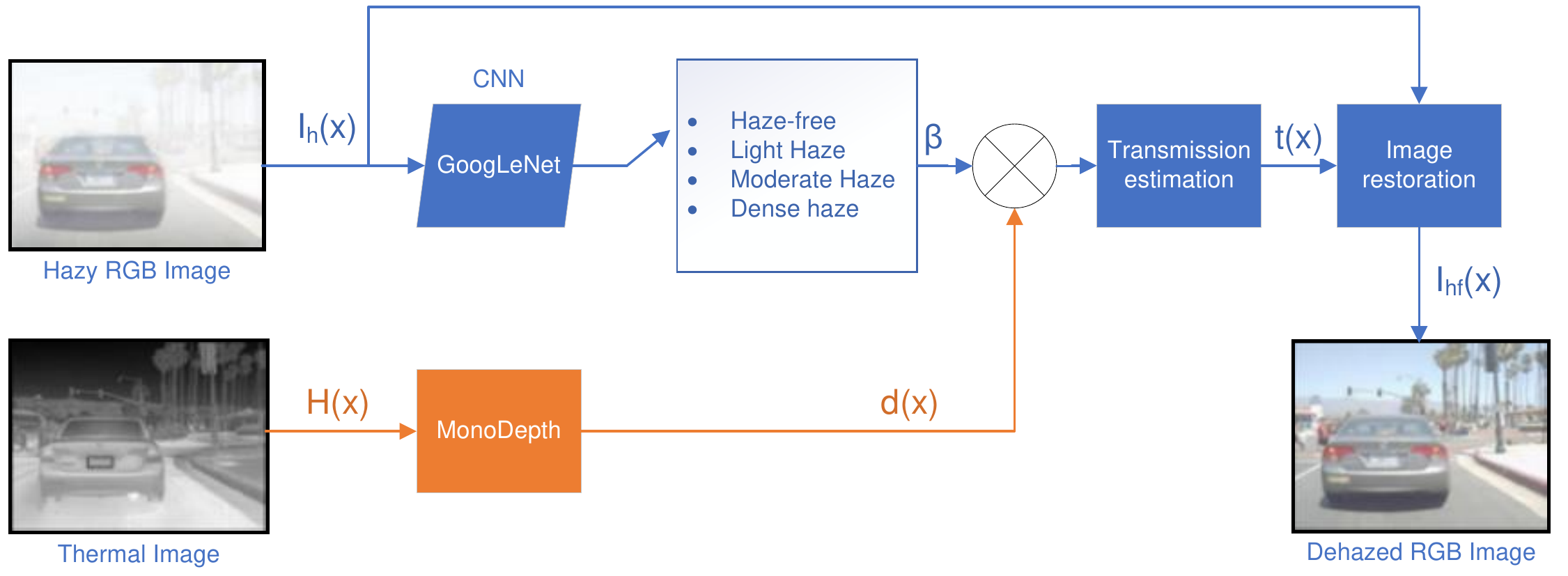}
    \caption{\hl{Depiction of the image dehazing process using GoogLeNet, a CNN-based classification model, to learn about weather conditions and select an appropriate atmospheric scattering coefficient based on the level of haze. The model performs depth estimation between the object and the camera using Monodepth and the thermal image. With the selected atmospheric scattering coefficient and depth information, a transmission map is estimated and a haze-free image is produced.}\cite{RN42}}
    \label{fig:RN42 F1 DN-RTD}
\end{figure*}
In \cite{RN93}, the authors utilised stereo vision and trained a neural network for disparity estimation to generate depth data. They also applied semantic segmentation, further discussed in section \ref{subsub:Semantic Segmentation}, using depth and RGB data to define the ROI for extracting thermal data and producing a 3D reconstruction for post-processing. Meanwhile, \cite{RN67} identified a region of interest(ROI) in the RGB modality also by segmentation but did this by classic methodologies not involving neural networks and extracting the thermal data by applying the ROI to the aligned thermal modality. In certain applications, such as those previously mentioned, the actual temperature values are relevant. However, in other studies like \cite{RN52}, the focus was on using the visual information derived from the thermal image rather than the actual temperature values. In these cases, transformations like stretching the brightness histogram values are applied to enhance the contrast, and additional denoising techniques are used to improve the image quality.

Unlike the previously discussed studies, the authors of \cite{RN71} configured a system in which ROIs are identified in the RGB modality based on the facial landmark points detected using the CLM Face Tracker\cite{CLM}, and their coordinates are converted to thermal frame coordinates. Key regions of interest include the facial area, ocular and periocular areas, and nose area, and evaluated parameters include position, orientation, green colour component, depth (distance), and temperature. The average values of each variable are computed for each region of interest, and the relative positions and temperatures are computed with respect to the average values computed for the entire face. Finally, each computed value is logged to an individual stamped CSV file for post-experimental processing and analysis.

Similarly, in \cite{RN78}, face detection and extraction of landmark points from RGB images are accomplished by using the Dlib \cite{king2009dlib} machine learning toolkit based on histogram-of-oriented-gradient (HOG) features. The authors assumed that the target person does not move significantly between two consecutive frames and limited consecutive detection to the previously identified area to increase the processing speed. The facial ROIs in the thermal image are located using calibrated landmark points. The forehead centre is computed as the middle of the two eyebrow corner points, and the average temperature in the forehead area is taken as the body temperature. The mean temperatures in the nose and cheek areas are used for the measurement of the respective respiration and heartbeat rates through harmonic analysis. The dominant frequency in the temperature signal's spectrum is identified by Fast Fourier Transform (FFT), and then multiplied by 60 to obtain the respiration or heartbeat rate in cycles per minute.

For the purpose of thermal comfort of humans, the authors in \cite{RN73} used algorithms implemented in OpenCV\cite{opencv_library} for facial tracking, but unlike in \cite{RN78}, there was no guarantee that a face faces the camera why the thermal images used for facial skin temperature measurements contain various types of noise, such as false detection of background as faces and interference from high-temperature objects in the environment, which are represented as sudden spikes in measurements. To remove such noise, the median filter was applied before data analysis. Unlike previous studies, that segmented the frontal face into several regions and collected skin temperature from each region, this study used global skin temperature features, including the highest, lowest, first quartile, third quartile, and average temperature measurements of all pixels in the detected facial region. These features provide an overall description of the distribution of skin temperature over a detected face, including both frontal and profile faces.

The authors in \cite{RN46} adopted a different approach for processing aligned modalities. They applied background substitution and evaluated the size of connected pixel areas from the delta image to determine whether a living being was detected or not. This study fused these regions of thermal and depth data at different levels to determine the optimal result. The study did not find any significant differences in the results based on the different fusion methods used. The evaluation resulted in an accuracy of 90.1\%. However, since the authors used their own data, no comparison with other methods was possible.

Numerous other studies \cite{RN74,RN71,RN75,RN102} have employed various detection methods to identify ROIs for extracting feature data to be used in decision systems or deep learning algorithms. For example, the average face temperature or the nostril area can be tracked to predict human behaviour. Further details on studies that extract data based on ROIs but process them further are presented in sections \ref{sub:Feature-Based Methods} and \ref{subsec:DeepLearningMethods}.
\section{Process Support}
\label{sub:ProcessSupport}
As briefly mentioned in the Fusion Stages\ref{subsec:FusionStages} section, there is also an indirect way of using a modality to improve the quality of the data of another modality. In \cite{RN42} the authors proposed a dehazing network with RGB and thermal depth (DN-RTD). To effectively remove haze, the DN-RTD dehazing network is designed to estimate $\beta$, the atmospheric scattering coefficient for the current atmospheric conditions, and $d(x)$, the depth between the camera and the object, using both RGB and thermal images. This network is shown in Fig. \ref{fig:RN42 F1 DN-RTD}.

In essence, the dehazing algorithm utilises GoogLeNet, a CNN-based classification model, to categorise captured hazy images $I_{h}\left(x\right)$ into four haze levels: haze-free, light haze, moderate haze, and dense hazy. The model then selects $\beta$ that corresponds to the classified weather condition. Additionally, the algorithm estimates depth information $d\left(x\right)$ from a thermal image $H\left(x\right)$ using Monodepth, rather than an RGB image. The transmission map $t\left(x\right)$, which expresses the level of atmospheric light transmission, is derived from an Equation using the estimated $\beta$ and $d\left(x\right)$. Finally, the clear image $I_{hf}\left(x\right)$ is extracted through the image restoration process. The authors then used two You Only Look Once (YOLO)\cite{RN118} detectors for both, the thermal and dehazed RGB image, and fused using late fusion. However, the dehazing process takes 659.1ms to compute why it is not suitable for real-time applications yet.
\section{3D Reconstruction}
\label{sub:3D Reconstruction}
3D thermal mapping reconstruction is a crucial application area for RGB-DT images. Based on the type of 3D reconstruction equipment used, 3D thermal mapping reconstruction methods can be categorised into five groups: RGB-D (ToF or Stereo Vision), Laser Scanning, binocular stereo-structured light encoding, Photogrammetry and Structure from Motion. 

The first depth camera employed to aid in 3D thermal mapping reconstruction was the Kinect v1, which has been used in various studies \cite{RN81,RN79,RN140}. The authors in \cite{RN96} developed a handheld 3D thermal mapping system using the Xtion Pro camera, and more recently, Kinect v2 and Intel RealSense\cite{RN70,RN72} have emerged as the most commonly used cameras for 3D thermal mapping reconstruction \cite{RN120,RN139,RN105}.

The most commonly used technique for large-scale 3D geometrical reconstruction is however Structure from Motion (SfM) \cite{ullman1979interpretation} which was utilised in \cite{RN98,RN94,RN97,RN39}. SfM-based 3D reconstruction approaches typically extract and track robust visual features (e.g. SIFT or SURF) on 2D images captured from different viewpoints and only work well under good illumination conditions (e.g. during daytime). Feature extraction and matching, which involves the detection of SIFT features, SURF features, ORB features, and AKAZE features, is a crucial part of the SFM algorithm. However, it only produces sparse 3D point clouds, and the generated 3D models lack absolute scale information, which is not ideal for thermal diagnosis applications.
To overcome these limitations, RGB-D-based 3D modelling approaches nowadays utilise depth sensors to acquire depth data of 3D objects/scenes from different viewpoints and apply 3D point cloud registration techniques, such as the iterative closest point algorithm, to align the current view with the global model \cite{newcombe2011kinectfusion,dai2017bundlefusion,whelan2015real,RN115}. Besides the better quality, it is also worth noting that binocular stereo-structured light, as used by \cite{RN138,sun2015three}, or time-of-flight depth sensors, can acquire 3D geometrical information in darkness.

Recently, the authors in \cite{RN138} introduced a fast and reliable 3D thermographic reconstruction method using stereo vision. The system features adjustable measurement fields and distances, based on the chosen optics for the cameras and projector. It can reach frame rates of up to 12.5 kHz for VIS cameras and 1 kHz for the LWIR camera at full resolution. By lowering the resolution, even higher frame rates can be attained.

Meanwhile, researchers in \cite{RN97} utilised terrestrial laser scanners (TLS) to acquire dense 3D point clouds, and temperature information obtained by an infrared camera is mapped onto 3D surfaces. To improve the mobility of 3D thermal imaging systems, a multi-sensor system consisting of a thermal camera and a depth sensor was built to generate 3D models with both visual and temperature information to be used for building energy efficiency monitoring \cite{RN79}.

Another method proposed was a thermal-guided 3D point cloud registration method (T-ICP) that improves the robustness and accuracy of 3D thermal reconstruction by integrating complementary information captured by thermal and depth sensors \cite{RN120}, but the method requires high computing resources to calculate several feature points. A set of experiments were performed to analyse how the key factors, such as sensing distance, specularity of the target, and scanning speed, affect the performance of high-fidelity 3D thermographic reconstruction. The authors in \cite{RN72} implemented a similar idea but the localisation method combines the ORB-SLAM2 with the thermal direct method, and the entire system runs on the Robot Operation System (ROS).

Based on the Thermal-guided Iterative Closest Point (T-ICP) algorithm presented in \cite{RN120}, the authors of \cite{RN105} developed a multi-sensor system that consists of a thermal camera, an RGB-D camera, and a digital projector. This method utilises an effective coarse-to-fine approach to enhance the robustness of pose estimation, allowing it to handle significant camera motion during large-scale thermal scanning processes. This system enables multimodal data acquisition, real-time 3D thermographic reconstruction, and spatial augmented reality through projection. 

\begin{figure}[ht]
    \centering
    \includegraphics[width=0.49\textwidth]{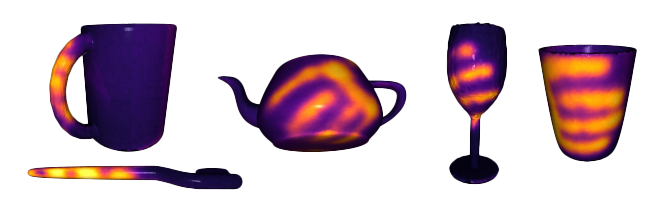}
    \caption{ Examples from ContactDB, constructed from multiple 2D thermal images of hand-object contact resulting from human grasps.\cite{RN139}}
    \label{fig:RN139 F1 Examples}
\end{figure}

A new dataset consisting of objects and their corresponding thermal imprints resulting from grasping was proposed in \cite{RN139}. To generate a coherent contact map of an object, the object is placed on a turntable which rotates as RGB-D and thermal images are captured from multiple viewpoints. The thermal images are texture-mapped onto the object's 3D mesh using a data processing technique. The steps involved in this process include extracting corresponding turntable angle and RGB, depth, and thermal images at nine locations where the turntable pauses, converting the depth maps to point clouds, estimating the turntable plane and segmenting the object using white colour segmentation, estimating the full 6D pose of the object in the nine segmented point clouds using the Iterative Closest Point (ICP) algorithm implemented in PCL, obtaining a least squares estimate of the 3D circle described by the moving object using the object origins in the nine views, and interpolating the object poses for views that are unsuitable for the ICP step. Finally, the 3D mesh along with the nine pose estimates and thermal images are input into a colourmap optimisation algorithm, which is implemented in Open3D\cite{Open3D2023}, to minimise the photometric texture projection error and generate a mesh that is coherently textured with contact maps. Examples of the resulting contact maps are shown in Fig. \ref{fig:RN139 F1 Examples}.

\section{Manual Descriptor-based Methods}
\label{sub:Feature-Based Methods}
In contrast to deep learning methods where the feature extraction is done by the Neural Network(NN), like by convolutional layers in Convolutional Neural Networks (CNN), handcrafted descriptors are manually designed features extracted from the input data. These include histogram of oriented gradients (HOG), Histogram of Optical Flow (HOF), scale-invariant feature transform (SIFT), local binary patterns (LBP), histograms of thermal intensities, oriented gradients (HIOG), and others. These techniques are considered to be the traditional methods that have been mostly replaced by CNNs, and more recently by transformer networks, in modern detection pipelines\cite{RN42}. The authors in \cite{RN91} noted that the SIFT algorithm is robust and accurate for matching features in RGB images, but it only computes low-level features and cannot recognise high-level representations. 

\subsection{Visual Modality (V)}
\label{subsub:VisualModality}
The authors in \cite{RN89} employed a combination of HOG, HOF, and HIOG to extract features. HOGs and HOFs are used to extract features from the RGB and depth data, while HIOG is used for thermal data. These features are then combined into a single feature vector and fed into a Random Forest classifier for body segmentation. The classic implementation of HOG was used for the RGB data but with a lower descriptor dimension than the original by not overlapping the HOG blocks. No gamma correction was used for the gradient computations and the Sobel kernel was applied. This means that for each pixel, the gradient orientation is determined by looking at the dominant colour channel (red, green, or blue) of that pixel, and then quantising it into a histogram over each HOG-cell \cite{RN89}.

HOF is a feature extraction method used to obtain motion information from an image. It works by computing dense optical flow and describing the distribution of the resultant vectors. The optical flow vectors are computed using the luminaries information of image pairs with the Gunnar Farnebäck’s\cite{RN110} algorithm. In \cite{RN89}, the authors used the implementation in OpenCV\cite{opencv_library}. The resulting motion vectors are then masked and quantised to produce weighted votes for local motion based on their magnitude, taking into account only those motion vectors that fall inside the colour grids. The votes are locally accumulated into a v-bin histogram over each grid cell according to the signed (0°–360°) vector orientations. Unlike HOG, HOF uses signed optical flow as the orientation information provides more discriminative power \cite{RN89}.

Similarly \cite{RN68} also used histogram-based descriptors but to process the colour modality, the RGB image is converted to a normalised colour space denoted as $rgb$, where $r = R/(R+G+B)$, $g = G/(R+G+B)$, and $b = B/(R+G+B)$. The colour normalisation approach is used to eliminate the illumination information in order to achieve robustness against lighting variations. Due to the fact that two components are adequate for describing the normalised colour space, with $r+g+b=1$, a 2D histogram $H_C$ is computed using the pair $(r,g)$.

\subsection{Depth Modality (D)}
\label{subsub:DepthModality}
For depth, the authors in \cite{RN89} used Histogram of Oriented Depth Normals (HON) to describe points in a point cloud. The depth modality contains a depth-dense map that represents a planar image of pixels measuring depth values in millimetres. The intrinsic parameters of the depth sensor can be used to obtain the actual coordinates from this depth representation, which can be seen as a 3D point cloud structure. This new representation allows measuring actual Euclidean distances that reflect the real world. After converting the depth modality, the surface normals for each point in the point cloud are computed, and their distribution of angles is summarised in an $\alpha$-bin histogram. Then a histogram describing the distribution of the normal vectors' orientations is built. A normal vector is expressed in spherical coordinates using three parameters: the radius, the inclination $\theta$, and the azimuth $\varphi$. In this case, the radius is a constant value, so this parameter can be omitted. 

For $\theta$ and $\varphi$ the calculation of the cartesian-to-spherical coordinate transformation is:
\begin{equation}
\theta = \arctan \left(\frac{n_z}{n_y}\right), \varphi = \arccos \frac{\sqrt{n_y^2+n_z^2}}{n_x} 
\label{eq:RN89 8 cartesian-to-spherical}
\end{equation}
Thus, a 3D normal vector can be represented by a pair of angles ($\theta, \varphi$), and the depth description comprises two histograms for $\delta_\theta$-bin and $\delta_\varphi$-bin, which are L1-normalised and combined. These histograms describe the angular distributions of the surface normals on the body.

Similarly, \cite{RN68} used an approach where a 3D normal vector is computed for each data point by fitting a 3D plane to a pre-defined local neighbourhood. Using the corresponding polar angle $\theta$ and azimuthal angle $\phi$ information, a 2D histogram $H_D$ is computed.

The authors in \cite{RN80} used a Leg Detection method proposed in an earlier work \cite{martinez2009laser} which utilises a probabilistic leg pattern. The leg model is implemented as a sequence of maximum, minimum, maximum, minimum, and maximum values based on the laser readings, as in \cite{bellotto2010bank}. Various measures are defined, such as the distance between the legs and the distance between the legs and background based on these five points. Besides the laser, the depth data of an RGB-D sensor is used to detect a particular emergency vest of a person. After detecting the corners, the Lucas-Kanade method is used to calculate the optical flow. The optical flow is computed for each corner, and in each frame, the centroid of the corners is then extracted, providing an estimation of the target's position.
\subsection{Thermal Modality (T)}
\label{subsub:ThermalModality}
\cite{RN89} used the Histogram of Thermal Intensities and Oriented Gradients (HIOG) descriptor derived from the thermal cue. This descriptor is a concatenation of two histograms. The first histogram provides a summary of thermal intensities, which are distributed over the range [0, 255]. The second histogram represents the orientations of thermal gradients. These gradients are calculated by convolving a first derivative kernel in both directions and then binned into a histogram, with their magnitude serving as a weighting factor. The two histograms are L1-normalised and concatenated. For the intensities, $\alpha_i$ bins, and for the gradient orientations $\alpha_g$ bins are used.

In a similar way, but solely relying on summarising the distribution of thermal intensities, the authors in \cite{RN68} proposed a method to generate a one-dimensional histogram for the thermal modality by directly utilising the intensity values of the thermal image.

Meanwhile, \cite{RN80} proposed a method to generate a 32-dimensional vector from a thermal image, where each element of the vector corresponds to the estimated probability of a person being present in a particular column of the image. This approach was chosen as the used thermal sensor had a resolution of 32x31 pixels. The computation of the vector involves three steps: firstly, a likelihood of a pixel corresponding to a person is assigned based on the assumption that the temperature of a person follows a normal distribution with mean and standard deviation values of 36 and 2, respectively, which are determined from several thermal images of people. Secondly, the likelihood matrix is smoothed by convolving it with a Gaussian kernel of a width of five pixels. Finally, the maximum value in each column of the likelihood matrix is used to determine the corresponding element of the output vector. The computation is based on established techniques such as the Lucas-Kanade optical flow method and Gaussian smoothing.

\subsection{General Feature Extraction}
\label{subsub:General Feature Extraction}
The authors in \cite{RN102} studied the detection of pain levels in faces and used the same feature extraction for all three modalities as a descriptor that considers both, spatial and temporal domains. This is needed to capture the spatiotemporal phenomena of changes due to pain in a facial expression. To achieve this, a steerable separable spatiotemporal filter has been selected, which utilises the second derivative of a Gaussian filter and their corresponding Hilbert transforms to measure the orientation and level of energy in the 3D space of $x, y$, and $t$. The filter provides information on the spatial texture of the face through its spatial responses and the dynamic of the features such as velocity through its temporal responses. The filter is applied independently to all three modalities, and for each pixel, the energy is calculated and normalised to improve comparability in different facial expressions. Finally, to improve localisation, the normalised energy is weighted using histograms of directions, and pixel-based energies are combined into region-based energies. For each pixel, the energy is calculated by:
\begin{equation}\label{eq: RN102 EQ2}
E(x, y, t, \theta, \gamma)=\left[G_2(\theta, \gamma) * I(x, y, t)\right]^2 \quad
\end{equation}
The convolution operator '*' is used to denote the operation in which $(x, y, t)$ represents the pixel value located at position $(x, y)$ of the $t$th frame (temporal domain) in the aligned video sequence $I$. $E(x, y, t, \theta, \gamma)$ represents the energy released by this pixel in the direction $\theta$ and scale $\gamma$. To ensure that the obtained energy measure is comparable across different facial expressions, normalisation is performed using:
\begin{equation}\label{eq: RN102 EQ3}
\hat{E}(x, y, t, \theta, \gamma)=\frac{E(x, y, t, \theta, \gamma)}{\sum E\left(x, y, t, \theta_i, \gamma\right)+\epsilon},
\end{equation}
After considering all directions $\theta_i$, where $i$ considers all directions and $\epsilon$ is a small bias to prevent numerical instability when the overall estimated energy is too small, the normalised energy is weighted to improve localisation using the method proposed in \cite{cannons2013applicability}:
\begin{equation}\label{eq: RN102 EQ4}
\dot{E}(x, y, t, \theta, \gamma)=\hat{E}(x, y, t, \theta, \gamma) . z(x, y, t, \theta)
\end{equation}
where:
\begin{equation}\label{eq: RN102 EQ5}
z(x, y, t, \theta)=\left\{\begin{array}{lc}
1 & \sum_{\gamma_i} \hat{E}\left(x, y, t, \theta, \gamma_i\right)>Z_\theta \\
0 & \text { Otherwise }
\end{array}\right.
\end{equation}
The resulting weighted normalised energy obtained in equation \ref{eq: RN102 EQ4} assigns a value to each pixel based on the level of energy released by that pixel, corresponding to the chosen directions of $\theta=0,90,180$, and 270. To combine these pixel-based energies into region-based energies, the authors follow study \cite{irani2015pain}, by using their histograms of directions:
\begin{equation}\label{eq: RN102 EQ6}
H_{R_i}\left(t, \theta_i, \gamma\right)=\sum_{R_i} \dot{E}\left(x, y, t, \theta_i, \gamma\right),
\end{equation}
The histogram $H_{R_i}$ represents the directions of the $i$-th region of the face, where $i=1,2$ or 3, and is used to combine regional histograms that are directly related to each other during the pain process. This is necessary because the muscles return to their original locations after being moved due to pain. In accordance with \cite{irani2015pain}, two directions of up-down (UD) and left-right (LR) are used to combine these histograms. The directional histograms are obtained for each modality of RGB, depth, and thermal, and are subsequently used separately to determine the level of pain.
\subsection{Segmentation}
This section explores various methods of basic segmentation using multiple modalities from the reviewed studies.

In \cite{RN67} the authors isolate the abdominal region of newborns. The regions of interest are extracted using depth information, followed by the refinement of the human body area using the colour information to remove the background and isolate the individual. First, a dynamic depth threshold is applied to separate the body from the flat bedding surface. The distance threshold is automatically determined based on the histogram of the depth map and the first significant observed cluster according to the imaging conditions, which involve imaging the subject from above. The second step involves utilising a skin colour model that is encoded in the YCbCr space to improve the segmentation of exposed body regions from other objects in the field of view, such as probes, tubes, or clothes. The method includes multiple steps using Canny edge detection and polygonal approximation algorithms.
Then, an additional refinement step is introduced in the form of a skeleton recognition method based on the depth image. This method utilises depth data to recognise various skeleton points and describe different parts of the human body.

The authors in \cite{RN70} proposed a multimodal egocentric SLAM(Simultaneous Localisation and Mapping) system based on ORB-SLAM\cite{mur2015orb} which faces a significant challenge in segmenting the input frame into left-hand, right-hand, object in interaction, and static environment classes. This segmentation is crucial for two reasons: first, removing dynamic points from the input frame is essential for successful SLAM operation, and second, these labels provide the necessary structure for proper scene understanding. The semantic segmentation algorithm the authors proposed is based on priors for the hands, including their colour model, temperature, and shape. Hand location is also a prior for the object in interaction. The segmentation is performed in two steps, first segmenting the left and right hands and then the object in interaction. The right and left hands are distinguished using the prior that the right hand is at the right side of the image frame and the left hand is at the left side. CRF-based image segmentation is used to segment the hands, defining an energy minimisation problem:
\begin{equation}
\min _{\alpha_i^t} \sum_i U\left(\alpha_i^t, \mathbf{y}_i^t\right)+\sum_i \sum_{j \in \mathcal{N}(i)} V\left(\mathbf{y}_i^t, \mathbf{y}_j^t\right) 1\left[\alpha_i^t \neq \alpha_j^t\right]
\end{equation}

In this equation, $\alpha_i^t$ represents a binary value of 1 if pixel $i$ is classified as part of the hand at time $t$, and 0 otherwise. The neighbouring set of $i$ is represented by $\mathcal{N}(i)$, and the indicator function is represented by $1(\cdot)$. The concatenated vector of $\mathbf{z}, \mathbf{c}, d, \tau$ is represented by $\mathbf{y}$. The unary energy function, $U\left(\alpha_i^t, \mathbf{y}_i^t\right)$, expresses the likelihood of pixel $i$ being part of the hand, and it is a weighted combination of the probabilities of temperature $(T)$, colour $(C)$, hand-detector outputs $(S)$, and history over time $(H)$.

\begin{equation}
\begin{aligned}
& U\left(\alpha_i^t, \mathbf{y}_i^t\right)=w_T U^T\left(\alpha_i^t, \mathbf{y}_i^t\right)+w_C U^C\left(\alpha_i^t, \mathbf{y}_i^t\right) \\
& +w_S U^S\left(\alpha_i^t, \mathbf{y}_i^t\right)+w_H \sum_i U\left(\alpha_i^{t-1}, \mathbf{y}_i^{t-1}\right) e^{-\Delta\left(\mathbf{y}_i^t, \mathbf{y}_i^{t-1}\right)}
\end{aligned}
\end{equation}

where $\Delta(\cdot, \cdot)$ calculates the geodesic distance over RGB-thermal space between two voxels. $V(\cdot, \cdot)$ is a binary consistency term that is defined over neighbouring pixels and takes the following form:
\begin{equation}
V\left(\mathbf{y}_i^t, \mathbf{y}_j^t\right)=\exp \left(-\frac{\left|\mathbf{y}_i^t-\mathbf{y}_j^t\right|_2}{\gamma}\right)
\end{equation}

where $\gamma=\frac{1}{N} \sum_i \frac{1}{|\mathcal{N}(i)|} \sum_{j \in \mathcal{N}(i)}\left|\mathbf{y}_i^t-\mathbf{y}_j^t\right|_2$, and $N$ is the total number of pixels.

They defined each component of the unary energy as:
\begin{equation}
\begin{aligned}
U^T\left(\alpha_i^t, \mathbf{y}_i^t\right) & =\tau_i^t 1\left[\alpha_i^t=1\right]+\left(1-\tau_i^t\right) 1\left[\alpha_i^t=0\right] \\
U^C\left(\alpha_i^t, \mathbf{y}_i^t\right) & =p\left(\mathbf{c}_i^t \mid \alpha_i^t\right) \\
U^S\left(\alpha_i^t, z_i^t\right) & =\sum_{k \in \mathcal{H}} p_k e^{-\Delta\left(\mathbf{y}_i^t, \mathbf{y}_k\right)} 1\left[\alpha_i^t=1\right]
\end{aligned}
\end{equation}

Where the RGB-colour model $p(\mathbf{c}_i^t|\alpha_i^t)$ is represented using a Gaussian Mixture Model (GMM) with five components and is learned separately for the hand and static scene from training data. $\mathcal{H}$ is a collection of hand detections, where each detection is represented by a centroid $\mathbf{c}_k$ and a detection likelihood $p_k\cdot\mathbf{y}_k$ which includes colour, position, depth, and temperature of the centroid of the detected hand.

All components of this energy function can be computed using bi-linear filters in log-linear time and minimised using the min-cut/max-flow framework as explained in \cite{csener2014efficient}. The authors used the open-source code released by the authors of \cite{csener2014efficient} and the original paper provides further details.

After segmenting the hands, the process continues to segment the remaining part of the image into static and dynamic object components. The same energy minimisation framework is used, with an additional motion prior and the removal of the colour prior. The motion prior accounts for the disparity between the motion of the object in interaction and the camera motion, and is defined as:

\begin{equation}
U^M\left(\alpha_i^t, \mathbf{y}_i^t\right)=\rho\left(\left|\mathbf{z}_i^t-\mathbf{z}_{\pi\left(\mathbf{R}^t \mathbf{x}_i^t+\mathbf{t}^t\right)}^{t-1}\right|\right)
\end{equation}
In the above equation, $\rho$ denotes the Huber function, $\pi$ is the pinhole projection, $\mathbf{R}$ and $\mathbf{t}$ are the estimated camera pose, and $\mathbf{X}_i$ represents the 3D position of the $i^{th}$ point in homogeneous coordinates. Here, $\alpha_i$ is a binary variable, which is equal to 1 if the $i^{th}$ pixel belongs to the object in interaction and 0 otherwise. 
The tradeoff parameters $\omega_T,\omega_S,\omega_H,\omega_M$ were learned by cross validation.

\subsection{Fusion \& Evaluation}\label{subsub: Fusion Feature}
In this section, we examine the various approaches adopted for fusing the extracted features and the evaluation processes of the reviewed studies.

While \cite{RN89} applied a background removal algorithm based on V and D for segmentation purposes to define the Ground Truth(GT), the authors in \cite{RN41} used background removal based only on D for their body segmentation in a fixed camera set-up. The paper proposes a modality fusion method for head tracking using D and T information for fall detection. This approach uses a particle filter to estimate the head position based on both, the D and T data. A Silhouette is constructed based on D and basic body shape assumptions, while the thermal data is used to distinguish the head from the background. The fusion is performed by combining and weighting D and T based on their reliability. The authors evaluated 4 different models and concluded that the D and T data was improving the results. However, the method was limited to 8 FPS and the authors further concluded to use Deep Learning models for future refinements of this application. In \cite{RN89}, the authors evaluated uni-modal classification and a multi-modal fusion based on a Random Forest classifier to achieve a human body segmentation with the extracted features discussed in section \ref{subsub:DepthModality}, \ref{subsub:VisualModality}, and \ref{subsub:ThermalModality}. 

\begin{figure}[ht]
  \centering
    \includegraphics[width=0.48\textwidth]{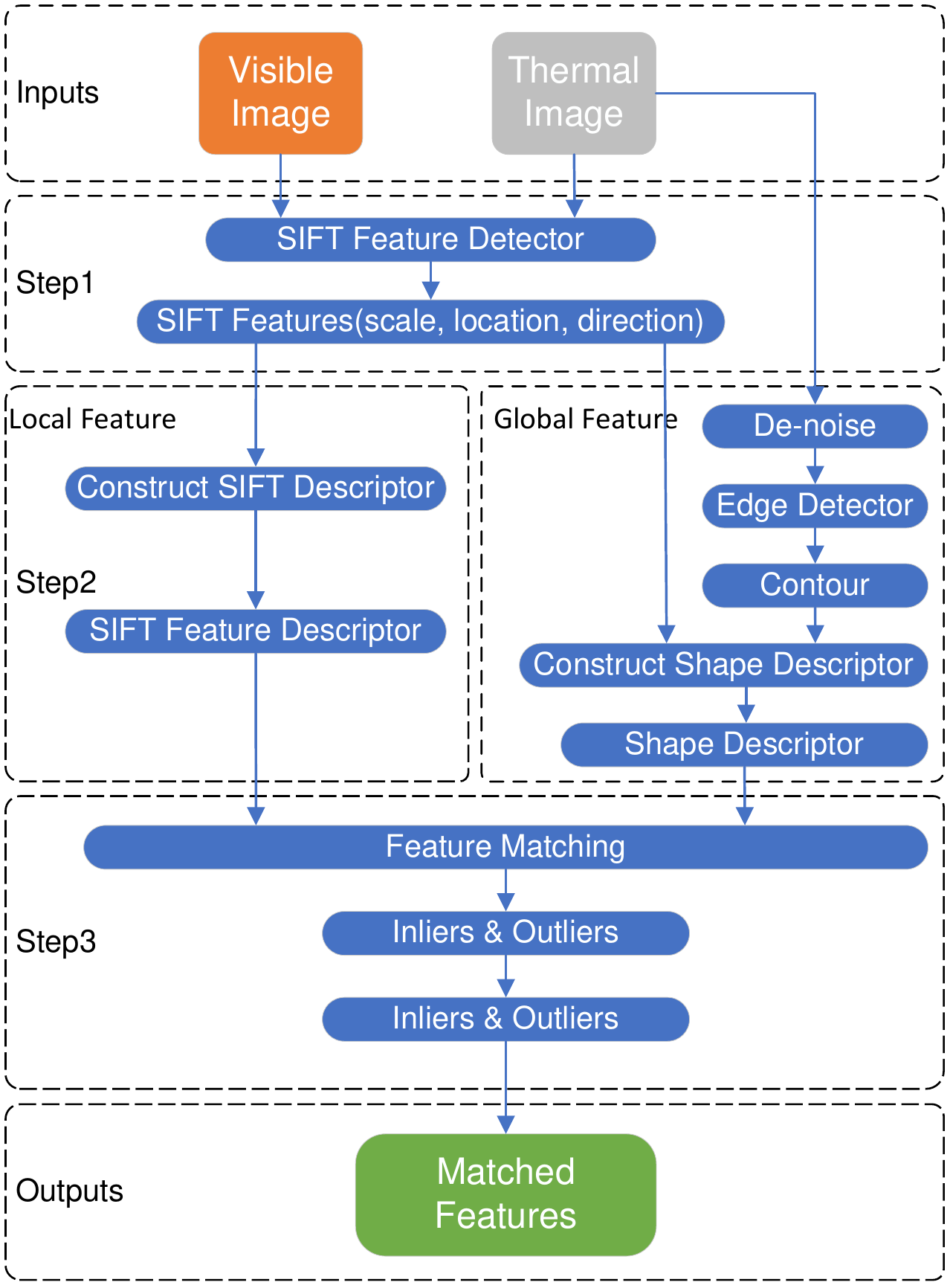}
    \caption{Structure of proposed feature matching algorithm.\cite{RN95}}
    \label{fig:RN95 F3 Feature matching}
\end{figure}

\cite{RN68} proposed a simple person-tracking algorithm that combines the parameters of the three modalities in a way that gives less weight to modalities where camouflaging occurs. The tracker's ability to resist significant radial motions was demonstrated using the Jaccard index. The target model was modelled using a single histogram for each data source and a histogram of 3D normals was used as the depth descriptor. This did not significantly improve the tracker's accuracy compared to the same approach without depth descriptors but the authors argued it could improve its robustness in more complicated sequences.

The People tracking system proposed in \cite{RN80} used four modalities and is based on a laser sensor, a thermal sensor and an RGB-D camera in a mobile setup. These sensors supply input to three detection units: Leg detection, Vest Detection, and Thermal Detection, which have been discussed previously. Once their individual likelihoods are calculated, the final likelihood is calculated using coefficients to weigh the three likelihoods. The inclusion of these coefficients enables assigning more importance to one of the information sources if desired, and the authors determined these values during the evaluation process.

The researchers in \cite{RN95} propose a feature-based registration method to register thermal and RGB-D images using the Shape Constrained SIFT Descriptor (SCSIFT). The registration process involves three steps: feature detection, feature description, and feature matching. In the first step, SIFT detector is applied to extract SIFT features from both visible and thermal images. In the second step, the proposed SCSIFT descriptor is constructed by combining the traditional SIFT descriptor with the shape descriptor extracted from the thermal image. In the third step, feature matching is performed by calculating the Euclidean distance between each shape descriptor vector and each SIFT descriptor vector, followed by normalisation and RANSAC to eliminate outlier matches. A detailed explanation of each this step can be found in the original study, Fig. \ref{fig:RN95 F3 Feature matching} depicts the proposed algorithm.

\noindent\textbf{Shape Feature Description:} Global descriptors to support local descriptors are added in multi-modality image feature matching. As the thermal image is noisy, anisotropic diffusion is applied for effective smoothing before edge extraction. The canny edge operator is used to extract edges, but contour-based methods alone are insufficient for correct feature matching. A circular template is generated around each feature point, with evenly plotted bins for edge point fitting. To describe the global position of the feature point and construct the shape descriptor, a spiral of Theodorus is applied to build the weighting function. The weighting of close region pixels is enhanced while the weighting of far region pixels is suppressed. The proposed SCSIFT descriptor is constructed from the entire image and adds a global shape constraint to the traditional SIFT descriptor which uses only local neighbourhood information.

\noindent\textbf{Feature Matching Scenario Based on SCSIFT Descriptor:} Normalisation is necessary before implementing RANSAC because the global shape descriptor vectors and local SIFT descriptor vectors are the statistical analysis of different information. For each feature $i$ in the source image $f_S^i$ with the descriptor denoted as $d_S^i$, the Euclidean distance of the global descriptor $d_{S(G)}^i$ and local descriptor $d_{S(L)}^i$ to all global and local features descriptors $D_{\text {ref }(G)}^{a l l}$ and $D_{\text {ref(L) }}^{a l l}$ in the reference image are calculated respectively, denoted as set $\boldsymbol{E}_G^i$ and set $\boldsymbol{E}_L^i$ shown in Eq. \ref{eq:RN95 7} and \ref{eq:RN95 8}.

\begin{equation}\label{eq:RN95 7}
\boldsymbol{E}_{\boldsymbol{G}}^i={\sqrt{\left(D_{\text {ref }(G)}^{a l l}-d_{S(G)}^i\right)^2}}^2 \\
\end{equation}

\begin{equation}\label{eq:RN95 8}
\boldsymbol{E}_L^i={\sqrt{\left(D_{\text {ref(L) }}^{a l l}-d_{S(l)}^i\right)^2}}^2 \\
\end{equation}

The ratio of the maximum value of sets $\boldsymbol{E}_G^i$ and $\boldsymbol{E}_L^i$ in Eq. \ref{eq:RN95 9} represents the scaling factor $S^i$. As the process of calculating Euclidean distances for descriptors is already an integral part of feature matching, this normalisation step does not add to the computational complexity.
\begin{equation}\label{eq:RN95 9}
S^i=\frac{\max \left(E_L^i\right)}{\max \left(E_G^i\right)}
\end{equation}
And the normalisation process is done by the Eq. \ref{eq:RN95 10}
\begin{equation}\label{eq:RN95 10}
\boldsymbol{E}^i=E_G^i \cup \frac{E_L^i}{S^i}
\end{equation}
The unified distance set $\boldsymbol{E}^i$ is used to determine the most likely match for $f_S^i$ to all features in the reference image, with the minimum value in $\boldsymbol{E}^i$ indicating the best match. Based on the maximum global and local distances, an appropriate scaling value is calculated for each feature to improve the matching accuracy. RANSAC is then employed to eliminate any outlier matches and refine the image transformation.

\section{Deep Learning-based Methods}
\label{subsec:DeepLearningMethods}
Deep learning-based approaches for multi-modal sensor fusion have gained increasing attention due to their ability to learn complex relationships between the different modalities and effectively fuse the information from multiple sensors. However, despite that images from multiple modalities can be beneficial in highlighting salient regions and providing more comprehensive information, they can also introduce interference between the different modalities\cite{RN38}. The application of Deep Neural Networks(DNN) can be categorised into semantic segmentation and object detection. In contrast to semantic segmentation, where multi-modal features are fused at various stages within the Fully Convolutional Network (FCN), object detection involves a wider range of network architectures and fusion variants. This diversity allows for greater flexibility and adaptability in addressing specific challenges related to object detection tasks\cite{RN112}. 

Convolutional Neural Networks (CNN) have long been the dominant architecture for image processing tasks. However, recent developments in applying transformer networks\cite{vaswani2017attention} to Computer Vision (CV), known as Vision Transformers (ViT)\cite{dosovitskiy2020vit}, have demonstrated high performance in segmentation, recognition, and detection tasks. These advances indicate the growing potential for transformer-based approaches in the field of CV.

Within the realm of multimodal object detection and segmentation, a considerable amount of research is directed towards autonomous driving applications, where the fusion of LiDAR point cloud data and RGB camera data is crucial. However, this paper focuses on RGB-D sensor data, which incorporates pre-aligned depth data or can be aligned using stereo calibration techniques. As a result, the methods for aligning point clouds with RGB data will not be discussed in this paper. However, once the RGB-D data is transformed into a point cloud, the succeeding methods for object recognition and detection can still be utilised. Although, papers using RGB stereo vision applying disparity prediction, briefly discussed in Section \ref{subsub:Disparity Prediction}, are included as this method generates similar data as RGB-D sensors.
\subsection{Disparity Prediction}
\label{subsub:Disparity Prediction}
To obtain depth data from a stereo image the disparity can either be computed, like in OpenCV which implements the block matching algorithm for calculating disparity with stereo calibration, or by training Neural Networks like AAnet (Atrous Adaptive Network)\cite{liu2019aanet}. AANet is a deep learning approach for stereo matching that can provide more accurate results than traditional methods such as block matching or simple disparity calculation. It can handle occlusions, textureless regions, and large baselines better than traditional methods. AANet is also more robust to lighting changes and can handle different camera configurations. Additionally, it can learn from large amounts of data, making it more adaptable to a wide range of scenarios. Overall, AANet provides a more flexible and accurate solution for stereo matching compared to traditional methods. Similar techniques can be applied to monocular vision to produce depth data as applied in \cite{RN42} using monodepth\cite{RN117}.

\cite{RN93} utilised AANet for disparity prediction of chicken images for feather damage analysis. The network extracts the down-sampled feature pyramid and constructs multi-scale 3D cost volumes\cite{gu2020cascade}. The cost volumes are then aggregated with six stacked Adaptive Aggregation Modules (AA Modules). Each AA Module consists of three Intra-Scale Aggregation (ISA) and a Cross-Scale Aggregation (CSA). The multi-scale disparity predictions are regressed by the soft argmin mechanism and hierarchically up-sampled and refined to the original resolution. The pre-trained AANet model for the Scene Flow dataset was used for direct inference on the dataset. The dataset was augmented by random colour augmentations and vertical flipping. The initial learning rate of the pre-trained AANet model was set to 0.001 and decreased by half at 400th, 600th, 800th and 900th epochs. Adam was used to optimise the parameters of the network to minimise the average loss of the model on the training data. The disparity range was from 0 to 192 pixels.

\begin{figure*}[ht]
    \centering
    \includegraphics[width=1\textwidth]{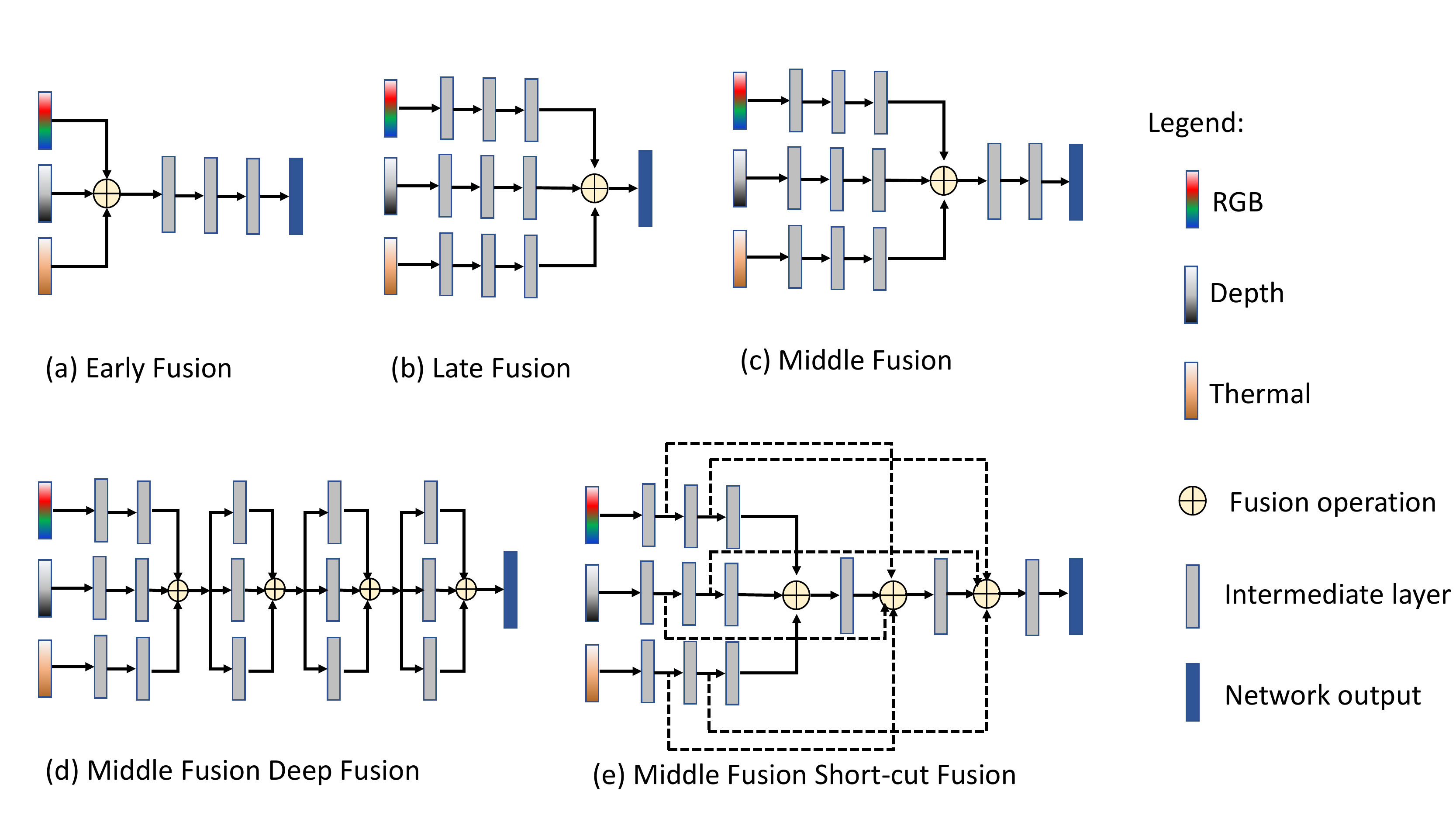}
    \caption{\hl{A depiction of architectures of different fusion schemes:} early fusion, late fusion, and various middle fusion techniques employed in deep learning environments based on \cite{RN112}.}
    \label{fig:000 Fusion Types}
\end{figure*}
\subsection{Fusion Methods (DL)}
\paragraph*{Data level fusion (DL)}
\label{para:Data Fusion DL}
In study \cite{RN112}, two primary advantages of early fusion were identified. Firstly, the network learns the joint features of multiple modalities at an early stage, allowing it to fully utilise the information present in the raw data. Secondly, early fusion has lower computational demands and requires less memory, as it processes multiple sensing modalities together. However, these benefits come at the cost of reduced model flexibility. For instance, when an input is replaced with a new sensing modality or the input channels are extended, the early fused network must be completely retrained which was also noted by authors in \cite{RN44}. This study created a two-channel image out of the two modalities for training a Faster R-CNN architecture with a ResNet-50 backbone. On the same dataset, IPHD, the authors in \cite{RN48} created a three-channel image by concatenating two duplicated thermal images and one depth image similar to a three-channel RGB image to make use of the ImageNet pre-trained weights for initialisation. They achieved similar results on the $AP_{50}$ metric compared to \cite{RN44}. In \cite{RN69} the authors fused the data into a 3-channel image, combining the grey channel of each modality per channel and applying different weights to the thermal and optical channels by applying the addWeighted method of OpenCV\cite{opencv_library}. This process involved a manual search to determine the optimal weighting scheme for the given data. To perform these unifications, the intensity values of the pixels from both images are multiplied by the desired weights and then added to compute the pixel intensity values of the resulting image. In this method, the thermal image pixel values are added to the optical image pixel values using weights ranging from 0.1 to 0.9 with a step size of 0.1.

The way a colour scheme is applied to translate the thermal to visual data plays a significant role, which is discussed in section \ref{subsub:LimModalities} and \ref{subsub:ObjectDetection}. Further, the method is sensitive to spatial-temporal data misalignment among sensors, which can be caused by calibration errors, different sampling rates, or sensor defects. This sensitivity further highlights the limitations of early fusion in certain scenarios\cite{RN112}.  

\paragraph*{Feature level fusion (DL)}
\label{para:Feature Fusion DL}
Middle fusion can be seen as a compromise between early and late fusion. By combining feature representations from different sensing modalities at intermediate layers, this approach allows the network to learn cross-modal information with varying feature representations and depths. Authors in \cite{RN112} argued that middle fusion is quite flexible, but that finding the "optimal" way to fuse intermediate layers for a specific network architecture can be challenging. \hl{This difficulty arises from the intricate interactions between features and the potentially vast array of possible fusion configurations. Nevertheless, merging feature representations from different sensing modalities at intermediate layers enables the network to learn cross-modalities with varying feature representations and depths. This fusion can occur at a specific layer only once or can be hierarchically fused, as depicted in Fig.} \ref{fig:000 Fusion Types}\hl{, such as through deep fusion or 'short-cut fusion'. Based on }\cite{RN112}\hl{, this figure further illustrates the intricate nature of middle fusion and the variety of approaches that can be taken to combine information from different sensing modalities.

Shortcut Fusion, as discussed in detail in the paper }\cite{ha2017mfnet}\hl{, is a technique employed in deep neural networks that involves creating additional pathways within the network. This allows early layers to directly contribute to later layers, aiming to combine the advantages of both early and late fusion. By utilising low-level feature fusion and high-level decision fusion, shortcut fusion has the potential to enhance accuracy. It preserves detailed information from earlier stages and incorporates it into the final decision stages. However, it's worth noting that this method may increase the complexity of the network, potentially requiring additional computational resources.

Deep Fusion, on the other hand, operates hierarchically at multiple levels within the network. This is beneficial in capturing intricate interactions between different modalities at an intermediate stage. The authors of} \cite{li2022deepfusion}\hl{ emphasise the importance of deep feature alignment for multi-modal object detection and how Deep Fusion improves detection accuracy and robustness against input corruptions and out-of-distribution data. Deep fusion allows for a more comprehensive understanding of the data, as it integrates information from various stages of processing. Furthermore, deep middle fusion is often favoured over late fusion due to its superior feature integration capabilities. By combining features at a deeper level, it can lead to a more robust and reliable model, thereby enhancing the overall performance.}

\hl{The complexity of middle fusion is further illustrated by the variety of techniques that can be implemented, as identified by the authors in }\cite{RN125}. These techniques include:
\begin{itemize}
    \item \textbf{Additive Fusion:} 
    Individual networks or branches process each sensing modality up to a designated intermediate layer.
    The feature maps from these intermediate layers are either added element-wise or concatenated.
    The resulting feature maps are further processed by the network to produce the final output.

    \item \textbf{Multiplicative Fusion:}
    Separate networks or branches handle each sensing modality up to a specific intermediate layer.
    The feature maps from these intermediate layers are multiplied element-wise.
    The combined feature maps undergo further processing within the network to generate the final output.

    \item \textbf{Skip Connections:}
    Separate networks or branches process each sensing modality.
    Feature representations from intermediate layers are combined via skip connections.
    The combined feature maps are further processed by the network to produce the final output.
\end{itemize}

\hl{Drawing on the insights from the authors of} \cite{RN112} and \cite{RN125}\hl{, it can be stated that Shortcut Fusion and Deep Fusion serve as overarching frameworks for integrating features from diverse modalities. Within these frameworks, specific techniques such as Additive Fusion, Multiplicative Fusion, and Skip Connections can be employed. Each of these methods brings its own unique strengths and potential challenges to the table. Additive Fusion, Multiplicative Fusion, and Skip Connections are distinct techniques that can be utilised to realise these fusion strategies. The choice between them depends on the specific requirements of the task at hand.}

In \cite{RN74}, CNNs were used for feature extraction on the visual input of two modalities. In addition, hand-crafted feature extraction was applied to the thermal data for fusing these features in a Support Vector Machine (SVM) model. In their study, the authors asserted that face temperature variation contains significant differences between different actions, and thus can enhance the accuracy of activity recognition. They utilised synchronised thermal images to extract the face temperature variation of participants while they performed the actions. To achieve this, they manually selected the face region in the first frame and tracked it across frames using a KCF tracker\cite{henriques2014high}. Outliers in face temperature were removed when the movement was sudden or when the person was partially out of the frame. They divided the temperature values into 25 intervals and computed the average temperature for each interval. Then, they calculated the difference between every two consecutive intervals $(t_i - t_{i-1})$, resulting in 24 features that were added to the SVM model.

The authors in \cite{RN75} extracted regions of interest (ROI) on the face using the Dlib\cite{king2009dlib} library to obtain the mean and variance of the temperatures in the ROIs as thermal features. The gait data of lower limbs were combined with these features for emotion detection, as lower limbs have more repeatable movements than the upper body. Joint angles and angular velocities were chosen as the features to characterize gait, including eight gait features based on the angle and velocity of the knees and hip. Convolutional Neural Networks (CNN), Hidden Markov Models (HMM), Support Vector Machines (SVM), and Random Forest (RF) models were employed to train and test the gait and thermal data. CNN and HMM models were trained with time series, whereas SVM and RF models were trained with static features such as the Power Spectral Density (PSD) of time series and the average temperatures of thermal image time series.

\paragraph*{Decision level fusion (DL)}
Late fusion on the other hand offers high flexibility and modularity. When a new sensing modality is introduced, only the network associated with that modality needs to be trained, leaving the other networks unaffected. However, late fusion comes with drawbacks, such as increased computation costs and memory requirements. Additionally, it discards rich intermediate features that could be highly beneficial if fused, potentially limiting the performance of the overall system\cite{RN112}. Late fusion in DL is commonly realised by the application of diﬀerent versions of the non-maximum suppression algorithm (NMS) which works by first selecting the bounding box with the highest object detection score. Then, it compares the remaining bounding boxes and removes the ones that have a high degree of overlap as applied in \cite{RN42}. \cite{RN44} investigated the NMS method further and compared the Dual-NMS with the simple method and concluded that the Dual-NMS had a better performance. The Dual-NMS involves sorting two lists of detection boxes based on their confidence scores and collecting pairs from them. Similar to the basic NMS method, the boxes with the highest scores from each list are selected one by one and compared with the boxes from the other list. If a sufficient intersection over union (IOU) is found, the detection box is paired with the candidate with the highest score from the other list. The paired boxes are then merged into a single result, and the final detection box coordinates are updated through weighted averaging of the coordinates of the components\cite{RN44}.

A depiction of various fusion types is shown in Fig. \ref{fig:000 Fusion Types}.
\subsection{Fusion}
The authors in \cite{RN112} noted that they did not find definitive evidence that one fusion method would be superior to others based on their review of various methods using different stages. However, \cite{RN44} and \cite{RN48} conducted a comparison of the performance between the early middle and late fusion techniques and concluded that early fusion yielded better results in their use cases. These studies focused on the fusion of depth and thermal data for human detection, and it was discovered in \cite{RN44} that using only depth data did not produce satisfactory detection outcomes. While the late fusion approach was slightly superior to using only depth data, it was inferior to using only the thermal data in comparison. However, \cite{RN48} further argued that early fusion outperforms the other fusion methods in both the final detection results and computational complexity. Unlike late fusion, which only merges the detected boxes, early fusion enables cooperation between the depth and thermal information during feature extraction, allowing the model to extract and combine useful information from both modalities. It should be noted that while intermediate fusion also merges feature maps, the merging in early fusion is accomplished by a deep backbone network, leading to more effective cooperation between the depth and thermal information. In their study, they also demonstrate that the use of a Receptive Enhancement Module (REM) improves AP by 0.4, 0.9, and 2.3 at IOU thresholds of 0.25, 0.5, and 0.75, respectively. These findings suggest that the REM module enhances the accuracy of bounding box localisation but \cite{RN44} achieved slightly better scores without it, which could be due to the slightly different fusion and training approach. In \cite{RN48}, the authors used a ResNet-50 backbone that was initialised with pre-trained parameters from ImageNet. However, the REM module and box prediction module were trained from scratch. \cite{RN44} used the same backbone but models that were pre-trained using the Common Objects in Context dataset (COCO).

Besides constructing custom networks, some studies, such as \cite{RN59}, have utilised deep learning-based algorithms like OpenPose\cite{openpose} to detect the pose of human occupants in a vehicle. Based on the derived bounding boxes this study applied late fusion. However, because OpenPose can only be applied to visual and thermal data, the authors did not utilise depth data. Other studies like \cite{RN74} use CNNs for the feature extraction on the visual input of two modalities and apply a hand-crafted feature extraction for the thermal data to fuse these in a Support Vector Machine(SVM) model. However, they dropped the thermal feature in their experiments due to too much noise in the data.
\subsection{Semantic Segmentation}
\label{subsub:Semantic Segmentation}
Image semantic segmentation is a crucial task in computer vision, serving as an ideal perception solution for transforming image inputs into semantically meaningful regions and enabling pixel-wise dense scene understanding. Networks that rely solely on RGB information may face limitations in segmentation performance in complex environments or under challenging conditions. To enhance input information and improve performance, researchers have extensively explored multimodal sensor data fusion, which integrates additional data sources to provide a more comprehensive understanding of the scene. Various approaches have been proposed, such as FuseNet \cite{RN125}, which incorporates depth information, and HeatNet \cite{RN126}, which leverages thermal data for improved performance at night. Polarisation information has also been integrated into models, as seen in EAFNet \cite{RN135}. Event data has been utilised in dense-to-sparse fusion to capture dynamic context information and improve segmentation performance, as in ISSAFE \cite{zhang2021issafe}. Furthermore, there are specialised methods for RGB-D \cite{RN129,RN130,RN123}, RGB-T \cite{RN131,RN132,RN133,RN134} and RGB-P \cite{RN92} semantic segmentation\cite{RN92}. 

The authors in \cite{RN136} argue that recently, vision transformers\cite{dosovitskiy2020vit} have gained attention as they handle inputs as sequences and can acquire long-range correlations, providing a unified framework for diverse multi-modal tasks. But that multi-modal data often contain noisy measurements in different sensing modalities, such as low-quality distance estimation regions caused by limited effective depth ranges \cite{hu2019acnet} and that compared to existing multi-modal fusion modules based on Convolutional Neural Networks (CNNs), it is not yet clear whether vision transformers can lead to significant improvements in RGB-X, where X stands for a different modality than RGB, semantic segmentation. Importantly, while some previous works like \cite{hu2019acnet} and \cite{chen2020bi} use a simple global multi-modal interaction strategy, it may not generalise well across different sensing data combinations\cite{zhang2021abmdrnet}. This is why the authors in \cite{RN136} hypothesise that for RGB-X semantic segmentation with various supplements and uncertainties, comprehensive cross-modal interactions should be provided to fully exploit the potential of cross-modal complementary features why they propose CMX, a method designed to enhance semantic segmentation by incorporating diverse and complementary information from multiple modalities. CMX is a transformer-based cross-modal fusion framework that uses two streams to extract features from RGB images and the X-modality and includes a Cross-Modal Feature Rectification Module (CM-FRM) in each feature extraction stage to calibrate the feature of the current modality by combining the feature from the other modality. 
A Feature Fusion Module (FFM) is then used to mix the rectified feature pairs for the final semantic prediction. FFM includes a cross-attention mechanism, enabling the exchange of long-range contexts, and enhancing bi-modal features globally. 
Using a SegFormer-B2 backbone to visualise the segmentation results demonstrated that CMX improves the semantic segmentation of RGB-D data and identifies objects correctly, which are misclassified by the RGB-only model. For RGB-T segmentation, CMX provides clearer boundary distinctions between persons and unlabeled backgrounds in low illumination conditions. For RGB-P, CMX accurately segments specular glass areas, cars with polarisation cues, and pedestrians. For RGB-Event, CMX enhances the segmentation of moving objects. For RGB-LiDAR, CMX correctly segments the scene as compared to the RGB-only method. The results show that CMX is a suitable approach for multi-modal sensing combinations, providing robust semantic scene understanding\cite{RN136}.
The proposed CMX framework achieves state-of-the-art performances in different benchmarks but is limited to two simultaneous modalities at the time of writing.

Similar to \cite{RN136}, the authors in \cite{RN92} put the focus on developing a generalisable multimodal perception system for various image modalities with an attention-based fusion architecture for outdoor scene understanding called NLFNet. This network is designed to effectively address the challenges of object segmentation in various complex scenarios. The NLF (Non-Local Fusion) module, a key component of the network, is capable of adaptively extracting and fusing complementary information from different modal input images. It also leverages dependence information along with long-range contextual and positional priors to enhance the accuracy of semantic segmentation and applies a weighting mechanism based on a sigmoid activation function for fusing the modalities. By addressing these challenges, NLFNet aims to improve the performance of outdoor scene understanding across a range of conditions and input modalities. 
The network architecture is inspired by efficient networks such as SwiftNet \cite{orsic2019defense} and RFNet \cite{sun2020real}. NLFNet uses an encoder-decoder structure and adopts a ResNet-18 \cite{he2016deep} backbone for each of its two independent branches. The encoder extracts latent features from RGB and other modal images, which are then merged using fusion operations. The Spatial Pyramid Pool (SPP) module \cite{zhao2017pyramid}, \cite{sun2020real} is employed to expand effective receptive fields and generate feature maps with more global contextual information.

NLFNet incorporates efficient upsampling modules from SwiftNet \cite{orsic2019defense} and merges RGB branch information through skip connections, improving segmentation accuracy. The Non-Local Fusion (NLF) module, inspired by Non-Local block \cite{wang2018non} and NANet \cite{RN130}, integrates complementary information from RGB and other branches for the multi-level fusion of feature maps. The NLF module consists of two sub-modules: the Spatial Dependency Module (SDM) and the Channel Dependency Module (CDM).

The SDM establishes long-range contextual dependency between RGB and other modal branches in space, using global average pooling and convolutions to expand receptive fields. The CDM concatenates outputs from the SDM module along the channel dimension, obtaining a merged feature map, and performs global average pooling to obtain a squeezed feature map. It then adaptively transforms these embeddings into dependency weights via a sigmoid activation layer. This process establishes non-local contextual dependencies between different modalities and extracts nonlinear interactions between cross-modal channels.

The authors of the study demonstrate the effectiveness and generalisation ability of NLFNet across various multimodal sensor combinations. By conducting experiments with different sensor data, such as RGB-Depth, RGB-Polarisation, and RGB-Thermal images, they showcase the ability of NLFNet to handle diverse modalities and effectively fuse the complementary information. The results indicate that NLFNet is capable of providing accurate semantic segmentation in various challenging scenarios, proving its potential as a robust solution for outdoor scene understanding. But like CMX \cite{RN136}, the solution is bi-modal only.
\begin{figure*}[ht]
    \centering
    \includegraphics[width=0.9\textwidth]{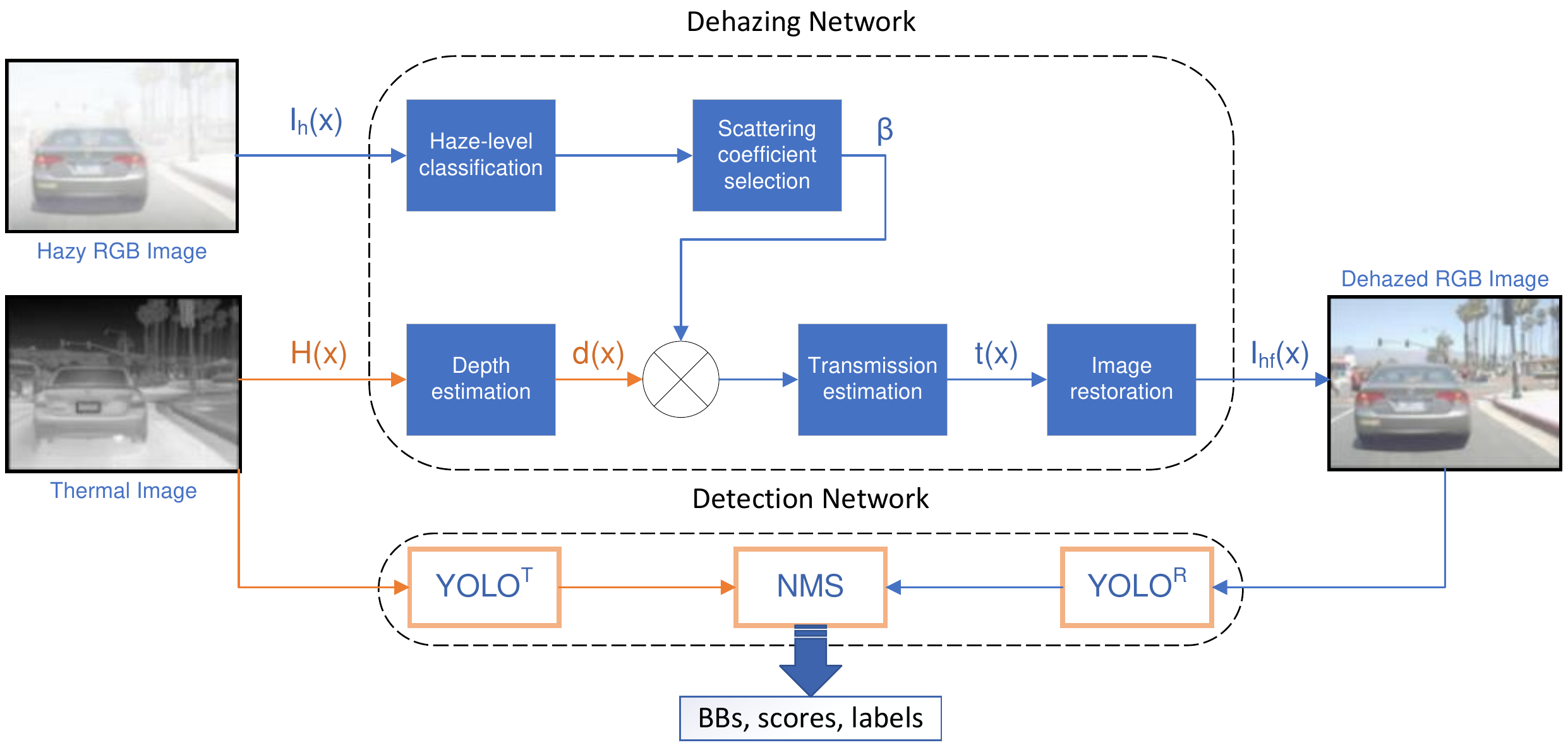}
    \caption{\hl{The overall architecture of the deep multimodal detection strategy. Object detection using YOLO from clear RGB images with rich colour information and thermal images with clear object bounding lines. The model detects the object with the highest probability through late fusion.}\cite{RN42}}
    \label{fig:RN42 F4 Architecture}
\end{figure*}

In \cite{RN93} applied the Residual Encoder-Decoder Network (RedNet) \cite{jiang2018rednet}, which is a high-performing semantic segmentation network proposed in \cite{RN123} that improves segmentation results by incorporating depth information into RGB signals. RedNet utilises an encoder-decoder network structure\cite{RN124} with residual blocks as building modules, as well as a pyramid supervision training scheme to optimise the network. The encoder structure includes two convolutional branches, one for RGB and one for depth, that have the same configuration except for the feature channel number of the convolution kernel, and feature fusion is achieved through element-wise summation. During training, the dataset was augmented and stochastic gradient descent (SGD) was used to optimise the network parameters with an initial learning rate of 0.002. The model is capable of segmenting target mask images from the background through inference. The authors augmented the training dataset by applying random scale and crop, followed by random hue, brightness, and saturation adjustment, which increased the dataset from 600 to 60,000 groups. Their model converged after approximately 100 epochs of training.

\subsection{Object Detection}
\label{subsub:ObjectDetection}
There are two main types of object detection algorithms that utilise convolutional neural networks (CNNs): two-stage detectors and single-stage detectors. The R-CNN family is a popular example of two-stage detectors, which typically use region-based methods. One such version is Faster R-CNN, used in \cite{RN44}, which introduced the region proposal network (RPN). The RPN can predict both the bounding box and the score at each position simultaneously, leading to a significant decrease in prediction time. An example of a popular single-stage network is YOLO\cite{RN118}. Most studies, \cite{RN69}\cite{RN42}\cite{RN70}, utilise these networks with slight adjustments and perform early data level fusion for RGB-DT or just RGB-D or RGB-T object detection. 

Only a single study, \cite{RN38}, focused on RGB-DT data fusion for Salient Object Detection (SOD) and implemented feature-level fusion with a CNN. In this work, the VGG16 classification network is used as a backbone for feature extraction. The tri-modal images are encoded separately using a three-stream encoding network, which extracts five-level features with varying resolutions. The authors proposed a hierarchical weighted suppress interference(HWSI) method to achieve an effective fusion of cross-modal information while also suppressing interference. The approach taken can be classified as a middle fusion with skip connections. This method involves assigning weights to each modality based on their importance for the given task and then using these weights to selectively suppress the interference introduced by each modality. By hierarchical weighting and selectively suppressing interference, the HWSI method can effectively fuse the cross-modal information but comes with a high computational cost which makes it less suitable for real-time applications. The HWSI method is composed of three distinct modules: the dual-modal attention fusion module (DMAFM), the triple-modal interactive weighting module (TMIWM), and the global attention-weighted fusion module (GAWFM). Each module is specifically designed to employ cross-modal information weighting to emphasize the salient regions and suppress interference effectively. 
The feature extraction is achieved by applying atrous convolutions with different dilation rates which can improve the performance of the network in tasks such as image segmentation and object detection. This approach, however, relies on the visual representation of the thermal and depth modality and some of the limitations that affect them are discussed in Section \ref{subsub:LimModalities}. The dataset created by the authors of \cite{RN38} limits the thermal and depth representation to 256 values and a dynamic colour AGC algorithm is applied to the thermal data, as shown in Fig. \ref{fig:000 VDT-2048 example} and no gain control is applied to the depth data as shown in Fig. \ref{fig:RN38 000 DepthImage18COMBO}. This can reduce the performance of object detection.

In contrast to the previous study, the authors of \cite{RN42} utilised late fusion with two separate YOLO\cite{RN118} models, following the thermal data dehazing process discussed in the Process Support\ref{sub:ProcessSupport} section. The overall system architecture is illustrated in Figure \ref{fig:RN42 F4 Architecture}. After completing the dehazing process, the resulting $I_{hf}(x)$ and $H(x)$ are fed into two YOLO models, denoted as $YOLO^R$ and $YOLO^T$, which use the image of their respective modality. Non-maximum suppression (NMS) is then employed to achieve late fusion. Their proposed model also allows for an RGB image with improved quality, as some of the haze removal can be performed using the haze level estimates. Moreover, by using late fusion and thermal images, the proposed model can process the rich colour and clear boundary information from both the RGB and thermal images simultaneously.

\cite{RN69} investigated using a pre-trained YOLOv4 network on the COCO dataset, as well as training YOLOv4 on their own dataset. They limited the scope to human detection only but investigated different ways of fusing the images in an early fusion as discussed here \ref{para:Data Fusion DL}. As their proposed data level fusion comes at a lower computational cost (average of 20ms. per frame) and the single-stage object detectors are fast, this system could be applied to real-time problems such as surveillance similar to study \cite{RN44}. However, neither study published data related to the inference speed, and while \cite{RN44} used a public dataset that was temporally aligned, \cite{RN69} did not discuss temporal sensor alignment and used a thermal camera that was limited to 8 FPS.

In \cite{RN70}, the authors employed all three modalities, RGB, depth, and thermal data, to detect hands using a YOLO-based object detection algorithm \cite{RN118} for real-time performance. Their analysis of the results suggested that using 2D bounding box detection with all three modalities led to higher accuracy compared to state-of-the-art model-based RGB-D hand pose detection algorithms. Additionally, the authors found that RGB and thermal data were the most crucial modalities for this task. To train the YOLO detector, pre-trained features on ImageNet \cite{deng2009imagenet} were used, and the annotated bounding boxes in the dataset were employed for training. Since pre-training is only available for RGB images, knowledge distillation \cite{RN122} was used to transfer pre-trained features to the thermal and depth modalities.

Overall, deep learning-based approaches for multi-modal sensor fusion have shown promising results in various applications, and their continued development is expected to significantly advance the capabilities of multi-modal sensing systems in the future.
\subsection{Presentation Attack Detection (PAD)}
\label{sub:PAD}

\begin{figure*}[htbp]
    \centering
    \includegraphics[width=0.9\textwidth]{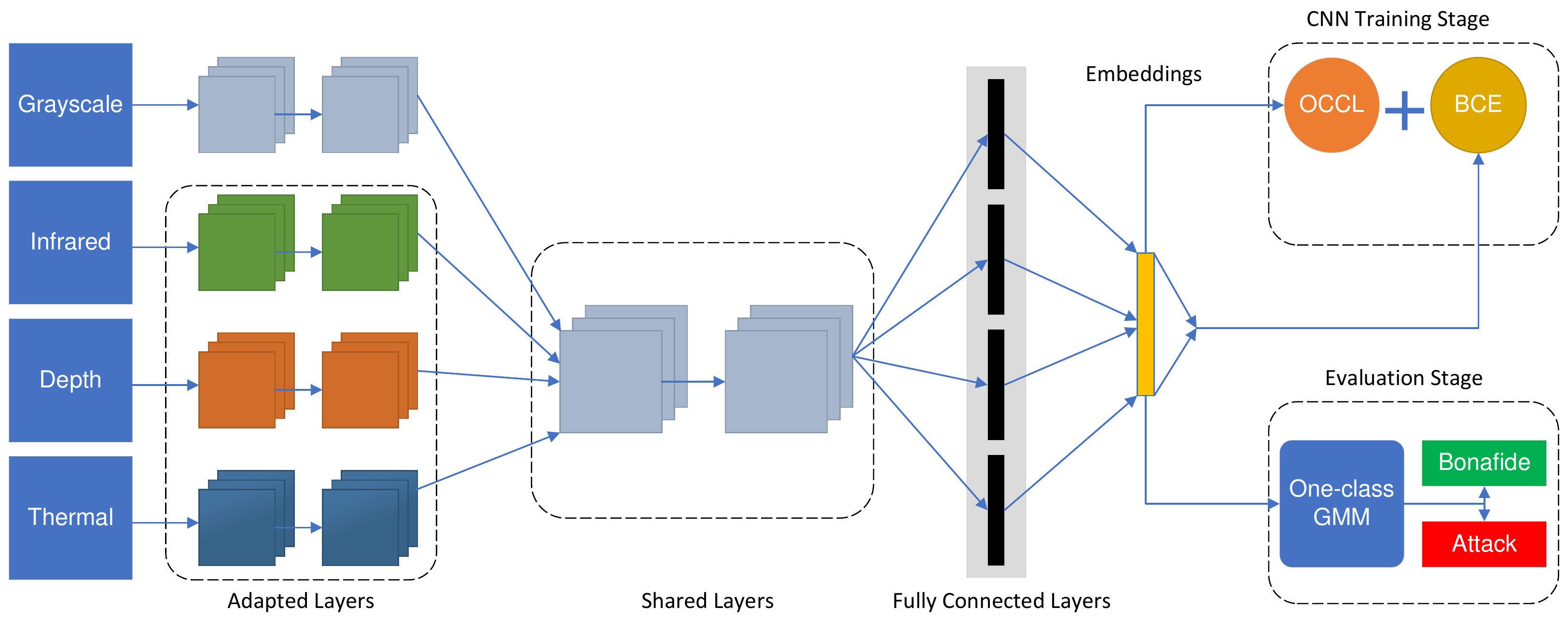}
    \caption{The CNN architecture is trained with two losses, \hl{the proposed One Class Contrastive Loss (OCCL) and Binary Cross Entropy (BCE),} and then used as a fixed feature extractor with frozen weights. The one-class Gaussian Mixture Model (GMM) is trained using the embeddings obtained from the bona fide class alone.\cite{RN141}}
    \label{fig:RN141 F4 SchematicDiagram}
\end{figure*}
Biometrics provides a secure and convenient method for access control. Among various biometric modalities, face biometrics is one of the most preferred due to its non-intrusive nature. Despite the high performance of systems in identifying individuals in many challenging datasets, they are still vulnerable to presentation attacks (PA). It was identified that PAD in visual spectra alone is insufficient for security-critical applications. This area has seen a lot of attention in recent years and PAD systems are another area where fused RGB-DT data was applied. A multi-channel PAD framework called the Multi-Channel Convolutional Neural Network (MCCNN) was proposed in \cite{RN119}. The MCCNN architecture is an extended version of the LightCNN model \cite{wu2018light} adapted specifically for multi-channel PAD tasks and was then also applied in \cite{RN141}. 
\begin{figure}[ht]
    \centering
    \includegraphics[width=0.48\textwidth]{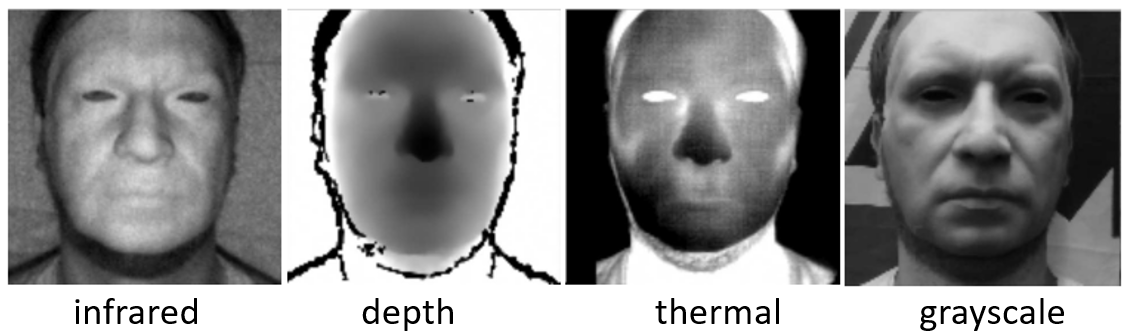}
    \caption{Preprocessed images resulting from a rigid mask attack \cite{RN141}.} 
    \label{fig:RN141 F3 PreprocessedImagesFromMaskAttack}
\end{figure}
The main idea behind the MCCNN architecture is to leverage the joint representation from multiple channels for PAD tasks, using a pre-trained face recognition network. In this approach, a pre-trained LightCNN face recognition model is extended to accept multiple channels, and the embeddings from all channels are concatenated. Two fully connected layers are added on top of this joint representation layer for the PAD task. The first fully connected layer has ten nodes, and the second layer has only one output node. The higher-level features in the LightCNN part are shared among all modalities. The advantage of this architecture is that only lower layer features, known as Domain Specific Units (DSUs) \cite{RN141}, and higher-level fully connected layers are adapted in the training phase. This approach has two main advantages: first, a smaller number of parameters since the high-level features are shared across modalities, and second, adapting only DSUs and the final fully connected layers reduces possible over-fitting since PAD databases are typically small in size. An optimal set of layers to be adapted was obtained empirically and was used in the baseline MCCNN and the proposed approach. Fig. \ref{fig:RN141 F3 PreprocessedImagesFromMaskAttack} shows a set of preprocessed images and Fig. \ref{fig:RN141 F4 SchematicDiagram} the Schematic diagram of the proposed system found in \cite{RN141}.

In \cite{RN141}, the authors proposed a framework that utilises a one-class classifier along with a novel loss function, which encourages the CNN to learn a compact yet discriminative representation for face images.

As part of study \cite{RN119}, a publicly available dataset called The Wide Multi-Channel Presentation Attack (WMCA) database, was released. \cite{RN142} made a similar dataset with higher quality and more modalities available which was called HQ-WCMA.
\begin{table*}[ht]
  \begin{minipage}[b]{1.0\textwidth}
\caption{Publicly available RGB-DT datasets.}   
\begin{tabular} {p{0.09\textwidth} p{0.3\textwidth} p{0.14\textwidth} p{0.12\textwidth} p{0.23\textwidth}@{}}
\hline 
Study & Name & Type & Modalities & Link\\
\hline 
\cite{RN38}  & VDT-2048 & General & RGB D T & \href{https://github.com/VDT-2048/VDT-Dataset}{VDT-Dataset}\\
\cite{RN89}  & VAP Trimodal People Seg.& People & RGB D T & \href{https://vap.aau.dk/vap-trimodal-people-segmentation-dataset}{VAP TPS-Dataset}\\
\cite{RN103}  & KAIST & Driving & RGB D\textsuperscript{*} T & \href{http://multispectral.kaist.ac.kr}{KAIST-Dataset}\\
\cite{RN119,RN141}  & WMCA & Faces/Masks & RGB D T & \href{https://www.idiap.ch/en/dataset/wmca}{WMCA-Dataset}\\
\cite{RN142}  & HQ-WMCA\cite{Mostaani_Idiap-RR-22-2020} & Faces/Masks & RGB D T & \href{https://www.idiap.ch/en/dataset/hq-wmca}{HQ-WMCA-Dataset}\\
\cite{RN45}  & TriModal Face Detection Dataset & Faces & RGB D T & \href{https://github.com/wiktormucha/tmfd_dataset}{TMFD-Dataset}\\
\hline
\end{tabular}
 \label{table:Datasets}
 \footnotesize\textsuperscript{*} Includes LiDAR and stereo RGB images.
  \end{minipage}
  \begin{minipage}[b]{1\textwidth}
\vspace{1em}
\centering
\caption{Publicly available Bi-Modal datasets used in studies.}   
\begin{tabular}  {p{0.09\textwidth} p{0.3\textwidth} p{0.14\textwidth} p{0.12\textwidth} p{0.23\textwidth}@{}}
\hline 
Study & Name & Type & Modalities & Link\\
\hline 
\cite{RN44,RN48} & IPHD\cite{clapes2020chalearn} & People & D T & \href{https://chalearnlap.cvc.uab.cat/dataset/34/description/}{IPHD-Dataset}\\

\cite{RN136} & NYU Depth V2\cite{silberman2012indoor} & Indoor & RGB D & \href{https://cs.nyu.edu/~silberman/datasets/nyu_depth_v2.html}{NYU Depth V2-Dataset}\\
\cite{RN136} & SUN-RGBD\cite{song2015sun} & Indoor & RGB D & \href{https://rgbd.cs.princeton.edu/}{SUN-RGBD-Dataset}\\
\cite{RN136} & Stanford2D3D\cite{armeni2017joint} & Indoor & RGB D & \href{http://buildingparser.stanford.edu/dataset.html}{Stanford2D3D-Dataset}\\
\cite{RN136} & ScanNetV2\cite{dai2017scannet} & Indoor & RGB D & \href{http://www.scan-net.org/}{ScanNetV2-Dataset}\\
\cite{RN136} & Cityscapes\cite{cordts2016cityscapes} & Driving & RGB D & \href{https://chalearnlap.cvc.uab.cat/dataset/34/description/}{Cityscapes-Dataset}\\
\cite{RN136} & MFNet\cite{ha2017mfnet} & Driving & RGB T & \href{http://www.mi.t.u-tokyo.ac.jp/static/projects/mil_multispectral}{MFNet-Dataset}\\
\cite{RN136} & EventScape\cite{gehrig2021combining} & Driving & RGB E & \href{https://rpg.ifi.uzh.ch/RAMNet.html}{EventScape-Dataset}\\
\cite{RN136} & KITTI-360\cite{liao2022kitti} & Driving & RGB L & \href{https://www.cvlibs.net/datasets/kitti-360/}{KITTI-360-Dataset}\\

\hline
\end{tabular}
 \label{table:BiDatasets}
\end{minipage}
\end{table*}
Three studies \cite{RN119, RN141, RN142} were co-written by some of the same authors who further developed their ideas in \cite{RN142}. The data was collected using a custom-made sensor suite that enabled the recording of both genuine faces and presentation attacks across five different image modalities, including RGB, NIR, SWIR, thermal, and depth. Moreover, four banks of six LED modules were employed for illumination, providing coverage in 10 different wavelengths ranging from 735nm to 1650nm, encompassing the NIR and SWIR spectra. Sequential switching of these infrared emitters, synchronised with camera exposure periods, yielded multi-spectral reflectivity data across the sample. These wavelengths were chosen to provide the best possible multi-spectral coverage given market availability, resulting in 14 different modalities in each recording, including four NIR and seven SWIR wavelengths. The cameras were co-registered using a calibration procedure, enabling the captured data to be aligned in each modality. Experimental results showed that the investigated CNN models with SWIR outperformed baselines when a wide variety of attacks was considered, with almost perfect detection of all impersonation attacks while maintaining a low BPCER. However, the generalisation ability of the models using SWIR data was assessed on a cross-database experiment, revealing a noticeable difference on bona fide attempts, suggesting future research directions.

The proposed database \ref{table:Datasets} and code for studies \cite{RN141}\footnote{https://gitlab.idiap.ch/bob/bob.paper.oneclass\_mccnn\_2019} and \cite{RN142}\footnote{https://gitlab.idiap.ch/bob/bob.paper.pad\_mccnns\_swirdiff} for reproducing the experiments are freely available for research purposes.

\section{Datasets}
The majority of studies examined in this paper faced a scarcity of publicly available datasets, leading them to develop their own. Although some studies claimed to make their datasets public, like the TriModal Face Detection dataset(TMFD) \cite{RN45}, it could not be found during the writing of this survey. \hl{However, subsequent to the preprint release of our paper, we were contacted by the authors of the TMFD, who have now made their dataset publicly available. This dataset is a comprehensive resource that encompasses a wide range of variations, including different numbers of people in the scene, various backgrounds and distances. The dataset is categorised into three separate groups based on the complexity and difficulty level of face detection.} Other studies, including \cite{RN38} and \cite{RN89}, created and made their datasets available \hl{and a list of all available tri-modal datasets can be found in Table }\ref{table:Datasets}. Public bi-modal datasets used by some reviewed papers are listed in Table \ref{table:BiDatasets}. As this paper is centred on RGB-DT tri-modal fusion, only datasets used by the studies included in this review are presented.

\section{Limitations}

\subsection{Sensors}
One of the limitations of using thermal cameras in conjunction with RGB-D cameras is the potential mismatch in their respective field of view (FOV) and focal length, which can restrict the effective distance between objects or subjects being monitored. This discrepancy can result in inconsistencies in the size, position, and orientation of objects in the captured images, which can impact the accuracy of object detection and tracking algorithms. Another thermal sensor limitation is the need for Non-Uniformity Correction (NUC). NUC compensates for inconsistencies in the sensor's response to temperature changes, which can lead to inaccuracies in temperature measurements. This correction is required periodically and involves a mechanical shutter operation that blocks the imaging sensor with a material of uniform temperature for a short time up to a second.

The authors in \cite{RN103} identified limitations with the temporal alignment in capturing images simultaneously using multiple devices. Despite the use of a signal generator to match the shutter times between devices, there can still be drift due to differences in exposure times. This can lead to an asynchronous phenomenon, especially during excessive movement of a vehicle or object.

It is important to note that temporal alignment is a critical factor in multi-camera systems, as it ensures that the images captured by different cameras are synchronised and can be properly used in applications such as 3D reconstruction or object detection.

\subsection{Modalities}
\label{subsub:LimModalities}

\begin{figure}[ht]
  \centering
  \includegraphics[width=0.99\linewidth]{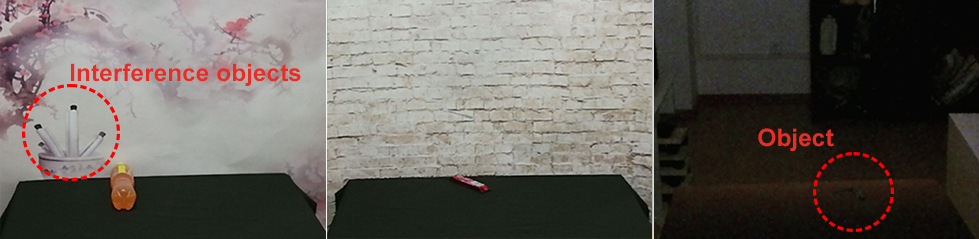}
  \bottomsubcaptionsABC
  \caption{RGB modality challenges. \subcaption{a}Similar appearance. \subcaption{b}Small salient object. \subcaption{c}Low illumination. \cite{RN38}}
  \label{fig:RN38 1 RGB challenge}
\end{figure}

In \cite{RN38}, the authors identified that visual perception systems that rely solely on RGB cameras face challenges such as:

\begin{enumerate}
\item The objects to be recognised in indoor environments are often small, numerous, dense, and vulnerable to background interference.
\item In low-light conditions, the ability to detect objects is greatly reduced, as illustrated in Fig. \ref{fig:RN38 1 RGB challenge}.
\end{enumerate}

\begin{figure}[ht]
  \centering
  \includegraphics[width=0.99\linewidth]{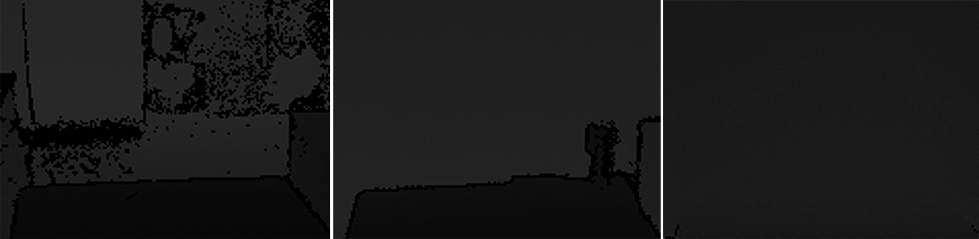}
  \bottomsubcaptionsABC
  \caption{Difﬁcult challenges of depth images. \subcaption{a}Background messy. \subcaption{b}Depth information is incomplete. \subcaption{c}Small salient objects. \cite{RN38}}
  \label{fig:RN38 F3 D challenge}
\end{figure}

\begin{figure*}[ht]
  \centering
  \includegraphics[width=0.98\linewidth]{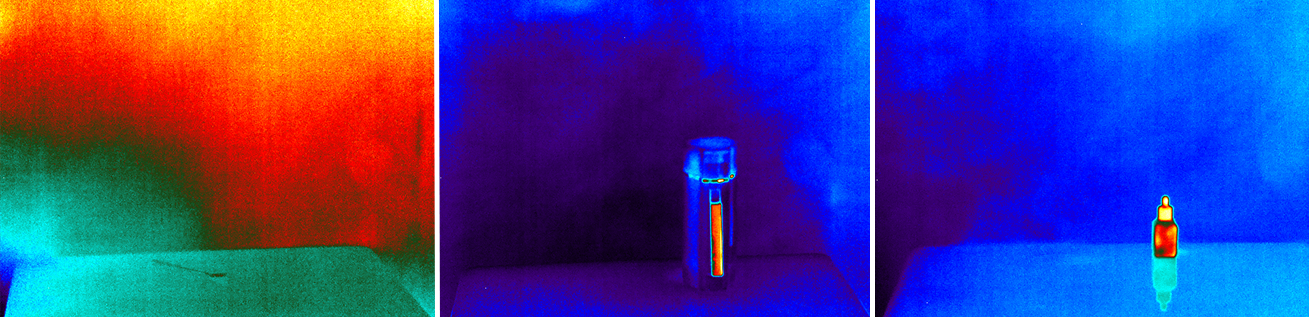}
  \bottomsubcaptionsABC
  \caption{\hl{Taken from the VDT-2048 dataset, demonstrating the three identified thermal challenges: }\subcaption{a} Thermal crossover, \subcaption{b} Thermal radiation dispersion, \subcaption{c} Heat reﬂection. \cite{RN38}}
  \label{fig:RN38 F4 T challenge}
\end{figure*}

To overcome the above problems, thermal and depth modalities can be introduced but despite that those sensors improve the detection of salient objects, they also introduce interference challenges and have their own individual challenges as can be seen in Fig. \ref{fig:RN38 F3 D challenge} and Fig. \ref{fig:RN38 F4 T challenge}.

Fig. \ref{fig:RN38 F3 D challenge}(a) illustrates that the background of the depth image without any salient objects is very cluttered, which can distract the detection of salient object detection algorithms. Also, the depth information of a salient object can be incomplete when there is no distance difference between it and the surrounding objects, or when the difference is minimal. Furthermore, depth sensing can still be challenging for detecting some small objects.

Thermal sensors also present several challenges that need to be addressed, including thermal crossover, thermal radiation dispersion, and heat reflection. Thermal crossover occurs when the temperature of a salient object is the same as that of a portion of the background, as illustrated in \ref{fig:RN38 F4 T challenge}(a), greatly increasing the difficulty of object detection. Fig. \ref{fig:RN38 F4 T challenge}(b) demonstrates an example of thermal radiation dispersion, where a portion of a salient object appears more salient than the rest of the object, causing interference to detection. Additionally, some objects exhibit heat reflection phenomena, as shown in \ref{fig:RN38 F4 T challenge}(c), which is another important interference that needs to be addressed.

Accurate assignment of thermal values, as identified in \cite{RN79}, requires careful attention due to the nature of thermal-infrared sensors. As regular NUCs are required, real-time systems must be able to cope with thermal data interruption. If the correctly measured temperature is of importance, it should be considered that the thermal value can also be affected by the incident angle between the sensor and radiation emitted from the surface. Minimising this angle is considered best practice. The authors of \cite{RN79} suggested three possible strategies to mitigate this:
\begin{itemize}
\item Perform Non-Uniformity Corrections (NUCs) more frequently, approximately every minute.
\item Disregard frames obtained from the camera while a NUC is in progress.
\item Assign temperatures only to rays with an incident angle of less than 30 degrees.
\end{itemize}
\section{Synthesis}
\label{Synthesis}
\hl{This section is intended to synthesise the key findings from our extensive review of the literature on the fusion of RGB-DT sensor modalities. It encapsulates the current state of the art, summarising the significant advancements and methodologies in this field across various applications. Before we transition into discussing the challenges, future work, and conclusion, this synthesis serves as a succinct recapitulation of the key points, including some insightful observations.}

The traditional approach for the geometric calibration of thermal cameras, using a printed chessboard and a flood lamp, was inaccurate and difficult to execute. Geometric masks with high thermal contrast were introduced as an alternative calibration pattern, and multi-material calibration boards made of two materials with different emissivities have been developed for cross-calibrating thermal and visual modalities which have proven to be reliable and accurate. The registration of the modalities is applied based on the processing requirements, with offline approaches utilising computationally intensive feature-point matching algorithms. One widely used technique, especially for large-scale 3D reconstruction, is the Bundle Block Adjustment (BBA) with current advancements and improvements to the basic approach, such as using more advanced optimisation algorithms (e.g. Levenberg-Marquardt\cite{more2006levenberg}, Gauss-Newton\cite{foresee1997gauss}) \cite{RN72}. 
In real-time 3D reconstruction, thermal data was added to the back-end of SLAM systems to enhance the robustness under unstable illumination environments and research in this area focuses on improving real-time performance by further reducing computational time and offering better model quality.
In contrast, real-time processing for semantic segmentation and object detection requires performing geometric image rectification and alignment as a preprocessing step to ensure correct feature extraction. While studies for multi-modal semantic segmentation of RGB, depth and thermal data based on recent transformer networks were found \cite{RN136,RN92}, those works only processed two modalities at a time and future research is aimed at processing more modalities simultaneously. However, the two studies demonstrate how to fuse bi-modal in real-time successfully and \cite{RN92} used an adaptive weighting of modalities with a sigmoid activation layer to limit interference.
The majority of the reviewed papers on multi-modal object detection utilised early fusion to generate a fused 8-bit three-channel image. Common object detectors such as YOLO were employed for the detection task, and some studies incorporated late fusion with the bounding boxes acquired from individual streams. A single study, \cite{RN38}, was identified that fused all three modalities (RGB, depth, and thermal) in a neural network using VGG16 as the feature extraction backbone. The study incorporated an interference suppression module to weigh the feature information across different modalities, mitigating interference from a single modality and compensating for potential information gaps in certain modalities. However, the computational requirements of this approach rendered it unsuitable for real-time processing. This study additionally created a publicly available, generic tri-modal dataset. Apart from this dataset, there are only three other RGB-DT datasets, which are specialised in presentation attack detection(PAD) and human detection applications. Another area that attracted a lot of attention is PAD where researchers focused on the problem of generalisation of the system. The challenge was addressed in \cite{RN141} by building upon the Multi-Channel CNN (MCCNN) originally proposed in \cite{RN119}. The authors developed a one-class classifier framework that employs learned features and a new loss function. This innovative loss function compels the CNN to acquire a concise and discriminative representation of face images, which enhances the overall performance even when used with the RGB modality alone. The authors showcased that their CNN method surpasses existing state-of-the-art feature-based techniques, while future research will focus on addressing the issue of potential attackers attempting to impersonate others.
\section{Challenges and future work}\label{Challenges}

\subsection{Data Fusion}
When it comes to fusing different modalities, one of the main challenges is sensor calibration and registration, especially when the sensors have different fields of view (FOV) that can cause parallax. To address this, techniques such as geometric calibration and image registration can be used to align the data from different sensors and reduce the effects of parallax. However, these techniques can be complex and time-consuming and still not produce perfect results, which can affect the accuracy of the fused data. Aside from employing software solutions to rectify parallax, another alternative involves using a beam-splitter, which enables two cameras to view the scene from the same point and utilise similar lenses to minimise parallax effects. Eliminating misalignment between modalities is vital in early fusion, which is why sensor calibration and registration continue to be significant research topics in this domain.
\begin{figure}[ht]
    \centering
    \includegraphics[width=0.48\textwidth]{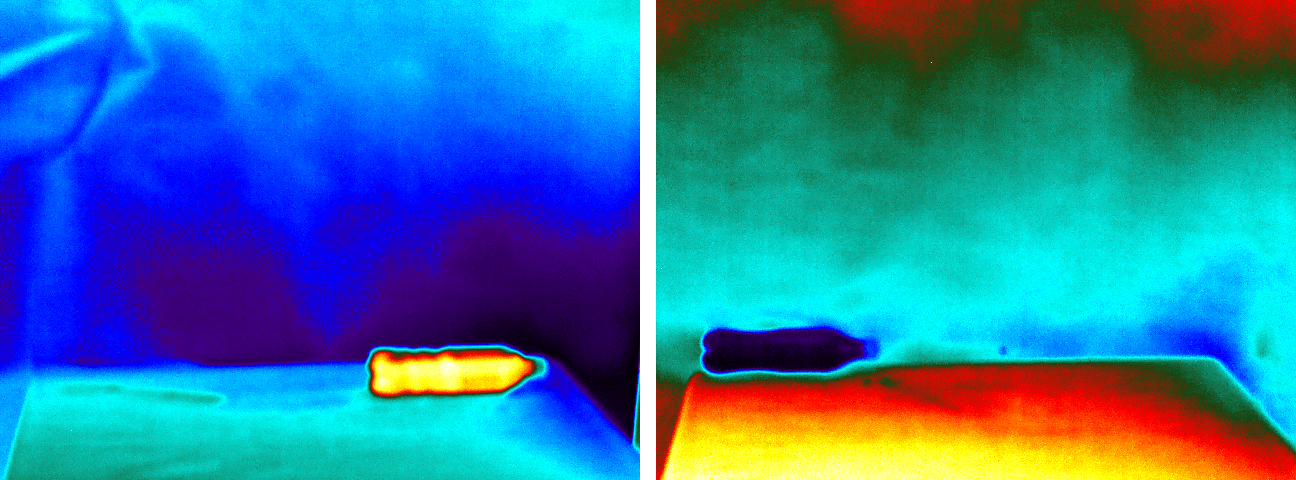}
    \caption{The image on the left shows a hot bottle on a table and the image on the right shows a cold bottle on the same table. Taken from the VDT-2048 dataset demonstrating the AGC colour shift of the same object(table) due to the application of a dynamic colour range based on the global minimum and maximum temperature in the frame.}
    \label{fig:000 VDT-2048 example}
\end{figure}

\begin{figure}[ht]
    \centering
    \includegraphics[width=0.48\textwidth]{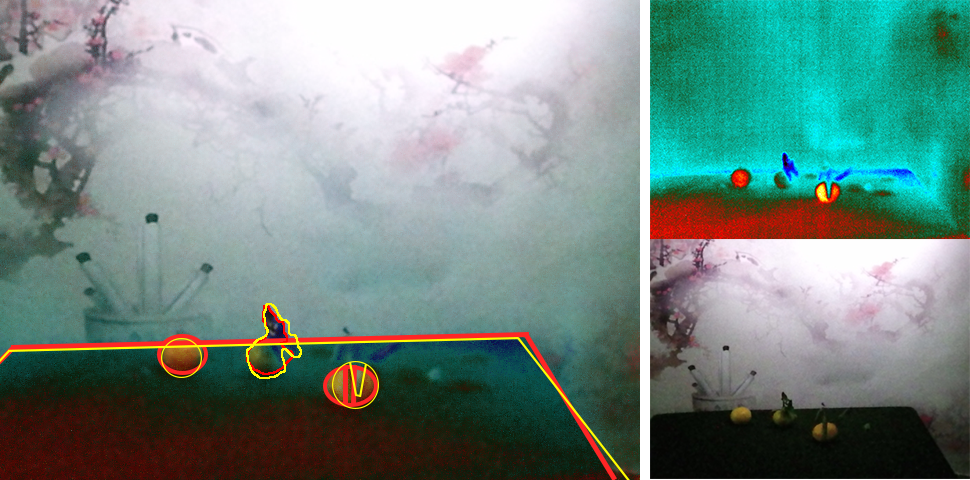}
    \caption{Taken from the VDT-2048 dataset, the image on the left shows an overlay of Visual(V) and Thermal(T) with some objects in T outlined in yellow and the same objects outlined in red in V. A parallax and distortion between the modalities can be observed. The images on the top right and on the bottom right are T and V respectively.}
    \label{fig:000 VDT-2048 distortion parallax}
\end{figure}
\subsection{Thermal Data}
The mapping of colours for the display of thermal data is a crucial element for systems that utilise thermal data in a visual form. However, there appears to be a shortage of discussion on this topic in the reviewed literature.
Most cameras apply automatic gain control (AGC) which is based on the lowest and highest temperature at any given time, causing the colours to shift. Besides that, grayscale or colour images limited to 256 values are being used but it is not specified over what range of temperatures it is used for e.g. when monitoring a range of -20 to 120°C, the resolution would be ~0.55°C. However, thermal cameras have a sensitivity expressed as Noise Equivalent Delta Temperature (NEdT), which can range from 0.020°C up to 0.075°C. It is crucial to consider both aspects when analysing thermal images as limiting the data to 8-bit discards detail. For example, the VDT-2048 dataset uses 256-colour thermal images with a dynamic thermal-to-colour range association, as shown in Fig.\ref{fig:000 VDT-2048 example}. The adaptive AGC algorithm may be appropriate for some applications; however, it could lead to difficulties if the intensity or colour information is essential for feature extraction or used for network training. Furthermore, the dataset exhibits some distortion and parallax between the visual and thermal modalities, as illustrated in Fig. \ref{fig:000 VDT-2048 distortion parallax}. Notably, the KAIST \cite{RN103} driving dataset features raw 14-bit thermal data; conversely, the pedestrian dataset only includes 8-bit data.
It is important to mention that while a 14-bit sensor can represent values up to 16,383, in environments with ambient temperatures around 20°C, the raw data captured falls within a narrow band of the full range. As a result, compression and contrast enhancement is crucial for encoding thermal images. However, it is essential to recognise that enhancement operations in thermal images can artificially distort the data, causing the loss of the physical correlation between the radiant flux from infrared radiation and pixel intensity \cite{RN143}. Besides the data preprocessing, authors in \cite{RN38} identified thermal crossover, thermal radiation dispersion, and heat reflection as challenges when processing thermal data. It is believed that a thorough preprocessing of thermal data and addressing the identified challenges in the field of multi-modal fusion constitutes a relevant future research direction.

\subsection{Depth Data}
\begin{figure}[ht]
    \centering
    \includegraphics[width=0.48\textwidth]{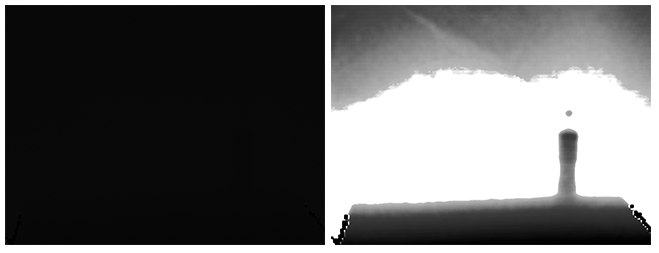}
    \caption{\hl{A comparative display of the original (left) and equalised (right) version of image 18 from the VTD-2048 dataset. The stark contrast between the two images accentuates the pivotal role of pre-processing in enhancing feature visibility, a critical step for the effective application of convolutional neural networks.}}
    \label{fig:RN38 000 DepthImage18COMBO}
\end{figure}

Similar to thermal data, depth data is initially captured with a 16-bit resolution, but it is later converted to an 8-bit format when used as a depth map unlike point clouds, which are usually generated from the raw values. This conversion from 16-bit to 8-bit can lead to a loss of depth resolution and information due to the reduction in detail. Although this process can result in significant information loss, no studies in the reviewed literature have addressed this issue. It is essential to conduct thorough preprocessing of this modality when using it in the form of an 8-bit depth map. While the dynamically applied AGC algorithm in thermal images could cause issues, applying no processing at all will result in a loss of details. Figure \ref{fig:RN38 000 DepthImage18COMBO} shows image 18 taken from the VDT-2048 dataset, in the original image on the left, it can be noted that visually almost nothing can be recognised as the observed depth is limited to a narrow band in the 16-bit data that was converted to 8-bit. On the right the same image with adjusted tonal balance by redistributing its brightness values. This is done by mapping the brightest and darkest pixel values in the image to white and black, respectively, and redistributing all the intermediate values evenly across the entire range.

\subsection{Datasets}
The research potential in the field of tri-modal RGB-DT object detection is currently limited due to the lack of publicly available datasets. Apart from the VDT-2048 dataset mentioned earlier, there are no other tri-modal datasets suitable for general object detection, highlighting the need for more comprehensive datasets to advance research in this area.

\subsection{Deep Learning}
Integrating multiple modalities can improve object detection or segmentation accuracy, but it also increases computational requirements and makes real-time processing difficult. 
In the literature, studies show that early and late fusion of data has a relatively small impact on real-time performance; however, the potential for more sophisticated enhancements remains to be a challenge. More complex processing techniques applied in middle fusion, as demonstrated in \cite{RN38} and \cite{RN42}, can result in frame rates dropping below 5 FPS. Even with the advances in Deep Neural Networks (DNNs) and hardware technology, achieving real-time object detection or segmentation with RGB-DT data still remains a challenge. While transformer-based architectures, as demonstrated in \cite{RN136}, and CNN-based architectures with non-local blocks, as demonstrated in \cite{RN92}, have shown promising results, they were limited to two concurrent modalities at the time of writing. Thus, further research is required to develop efficient and accurate fusion algorithms that can utilise three modalities and meet the requirements of real-time processing.

\subsection{PAD}
Currently, research in the field of PAD (Presentation Attack Detection) algorithms is centred on devising new methods capable of accurately detecting both known and unknown attacks. A major challenge for existing PAD algorithms is generalisation, as they often exhibit bias towards the training data. In \cite{RN142}, the authors recognised the SWIR (Short-Wave Infrared) spectrum, typically defined as light in the 0.9 – 1.7$\mu$m wavelength range but can also range from 0.7 – 2.5$\mu$m, as complementary and valuable. However, InGaAs (Indium Gallium Arsenide) sensor-based cameras for this spectrum are costly and reserved for specific applications. Meanwhile, in \cite{RN141} the researchers tackled the generalisation issue by building upon the Multi-Channel CNN (MCCNN) initially proposed in \cite{RN119}. Their CNN approach with an innovative loss function outperformed all other methods. Future research will concentrate on addressing the challenge of potential attackers attempting to impersonate others.
\section{Conclusion}\label{Conclusion}
This paper presents a comprehensive overview of the fusion between RGB-D and thermal modalities, exploring their applications, and the techniques employed. Over the past decade, there has been a surge of interest in fusing these modalities, demonstrating their considerable potential across diverse fields, including robotics, surveillance, medical imaging, and maintenance systems. Combining these modalities has proven to enhance the accuracy, robustness, and reliability of computer vision systems, contributing to the overall effectiveness of the technology. To systematically summarise the findings, a search strategy based on the PRISMA framework was used. The literature review has revealed several approaches for integrating RGB-D and thermal data, including feature-level fusion, decision-level fusion, and data-level fusion. Furthermore, the use of deep learning techniques has emerged as a popular approach for effectively combining RGB-D and thermal data, surpassing traditional feature-based approaches. Overall, the reviewed literature suggests that the fusion of RGB-D and thermal modalities holds great potential for enhancing the performance of computer vision systems in diverse applications and even creating new ones. 

It was observed that researchers have primarily focused on the higher-level architecture of neural networks when conducting sensor fusion with deep learning while overlooking the importance of preprocessing steps. While Visual Transformers have shown promising results in sensor fusion, no existing tri-modal RGB-DT fusion has been developed thus far. Therefore, further research is necessary to develop advanced fusion techniques that can enhance the accuracy and reliability of the results while operating in real-time, thereby unlocking the full potential of this approach and making it applicable for various practical applications. In conclusion, this study aims to serve as a supplementary resource for researchers in the field of RGB-DT sensor fusion, providing a robust foundation and guidance for ongoing investigation and advancements.

\bibliographystyle{IEEEtran}
\bibliography{bibliography} %

\begin{myBiographyNoPhoto}{Martin Brenner} received the B.Sc. in Information Technology and Computing from the Open University, Milton Keynes, UK, in 2008, and the M.Sc. degree in Computer Science from the Massey University, Auckland, NZ, in 2021. He is currently pursuing a Ph.D. degree at Massey University.
\end{myBiographyNoPhoto}

\begin{myBiographyNoPhoto}{Napoleon H. Reyes} received the B.Sc. degree in Physics, M.Sc. and Ph.D. (with High Distinction) degrees in Computer Science from De La Salle University, Manila, Philippines, in 1993, 1999 and 2004, respectively.  He is currently teaching at Massey University, Auckland, New Zealand. His research interests include evolving meta-level reasoning, intelligent systems and deep learning.
\end{myBiographyNoPhoto}

\begin{myBiographyNoPhoto}{Teo Susnjak} holds a Ph.D. in Computer Science with a focus on machine learning. He teaches data science topics at both undergraduate and postgraduate levels at Massey University. He has broad industry experience, having worked as a machine learning analyst, and he continues to apply his research to solving real-world problems using data-driven approaches.
\end{myBiographyNoPhoto}

\begin{myBiographyNoPhoto}{Andre L.C. Barczak} received the B.Eng. degree in mechanical engineering from Unicamp, Brazil, in 1987, the M.S. degree from Unicamp, Brazil, in 1996 and the Ph.D. degree in computer science from Massey University, New Zealand, in 2008. From 1989 to 1995, he was working as a system engineer with IBM. He was a research assistant with Unicamp from 1995 to 1998. In 1998 he emigrated to New Zealand and worked as a system engineer until 2002. Between 2002 and 2022, he worked as an academic staff with Massey University in computer science. Since 2023 he works at Bond University in Australia. He has published more than 70 technical articles in journals and conferences. His research interests include computer vision, machine learning, big data and optimisation algorithms.
\end{myBiographyNoPhoto}
\vfill
\end{document}